\documentclass[lettersize,journal]{IEEEtran}
\usepackage{amsmath,amsfonts}
\usepackage{algorithmic}
\usepackage{algorithm}
\usepackage{array}
\usepackage[caption=false,font=normalsize,labelfont=sf,textfont=sf]{subfig}
\usepackage{textcomp}
\usepackage{stfloats}
\usepackage{verbatim}
\usepackage{graphicx}
\usepackage{cite}
\usepackage[hyphens,spaces,obeyspaces]{url}
\usepackage{hyperref}

\usepackage{cleveref}
\usepackage{tabularx}

\usepackage{amsmath,amssymb,amsfonts}
\usepackage{afterpage}
\usepackage{pifont}
\usepackage{authblk}

\begin{document}

\title{Automatic Modulation Classification with Deep Neural Networks}

\author{Clayton~A.~Harper,
        Mitchell~A.~Thornton,
        and~Eric~C.~Larson
}

\affil{Darwin Deason Institute for Cyber Security\ \\
    \{caharper, mitch, eclarson\}@smu.edu
}






\maketitle

\begin{abstract}
    Automatic modulation classification is a desired feature in many modern software-defined radios. In recent years, a number of convolutional deep learning architectures have been proposed for automatically classifying the modulation used on observed signal bursts. However, a comprehensive analysis of these differing architectures and importance of each design element has not been carried out. Thus it is unclear what tradeoffs the differing designs of these convolutional neural networks might have. In this research, we investigate numerous architectures for automatic modulation classification and perform a comprehensive ablation study to investigate the impacts of varying hyperparameters and design elements on automatic modulation classification performance. We show that a new state of the art in performance can be achieved using a subset of the studied design elements. In particular, we show that a combination of dilated convolutions, statistics pooling, and squeeze-and-excitation units results in the strongest performing classifier. We further investigate this best performer according to various other criteria, including short signal bursts, common misclassifications, and performance across differing modulation categories and modes.   
\end{abstract}


\section{Introduction}
\label{sec:introduction}
    \IEEEPARstart{A}{utomatic} modulation classification (AMC) is of particular interest for radio frequency (RF) analysis and in modern software-defined radios to perform numerous tasks including ``spectrum interference monitoring, radio fault detection, dynamic spectrum access, opportunistic mesh networking, and numerous regulatory and defense applications''\cite{deepsig}.  Upon detection of an RF signal with unknown characteristics, AMC is a crucial initial procedure in order to demodulate the signal.  Efficient AMC allows for maximal usage of transmission mediums and can provide resilience in modern cognitive radios.  Systems capable of adaptive modulation schemes can monitor current channel conditions with AMC and adjust exercised modulation schemes to maximize usage across the transmission medium.
    
    Moreover, for receivers that have a versatile demodulation capability, AMC is a requisite task. The correct demodulation scheme must be applied to recover the modulated message within a detected signal. In systems where the modulation scheme is not known \textit{a priori}, AMC allows for efficient prediction of the employed modulation scheme.  Higher performing AMC can increase the throughput and accuracy of these systems; therefore, AMC is currently an important research topic in the fields of machine learning and communication systems, specifically for software-defined radios.

    Typical benchmarks are constructed on the premise that the AMC model must classify not only the mode of modulation (\textit{e.g.}, QAM), but the exact variant of that mode of modulation (\textit{e.g.}, 32QAM).  While many architectures have proven to be effective at high signal to noise ratios (SNRs), performance degrades significantly at lower SNRs that often occur in real-world applications.  Other works have investigated increasing classification performance at lower SNR levels through the use of SNR-specific modulation classifiers \cite{harper2021snr} and clustering based on SNR ranges \cite{soltani2019spectrum}.  
    To perform classification, a variety of signal features have been investigated.  Historically, AMC has relied upon statistical moments and higher order cumulants \cite{old_stat_moments, mod_class_cumulants, coop_cumulants} derived from the received signal.  Recent approaches \cite{asilomar_2020,deepsig_original,deepsig,skip_connections} use raw time-domain in-phase (I) and quadrature (Q) components as features to predict the modulation variant of a signal.  Further works have investigated additional features including I/Q constellation plots \cite{Peng, ConstellationDiagrams, GraphicConstellations}.
    
    After selecting the signal input features, machine learning models are used to determine statistical patterns in the data for the classification task.  Support vector machines, decision trees, and neural networks are commonly used classifiers for this application \cite{nn_example_1,tree_ex_1,deepsig,asilomar_2020,deepsig_original,skip_connections, soltani2019spectrum, Peng}.  Residual neural networks (ResNets), along with convolutional neural networks (CNNs), have been shown to achieve high classification performance for AMC \cite{deepsig_original, deepsig, asilomar_2020, skip_connections, Peng, soltani2019spectrum}.  Thus, deep learning based methods in AMC have become more prevalent due to their promising performance and their ability to generalize to large, complex datasets.
    
    \begin{figure*}[htbp]
        \centering
        \includegraphics[width=\textwidth]{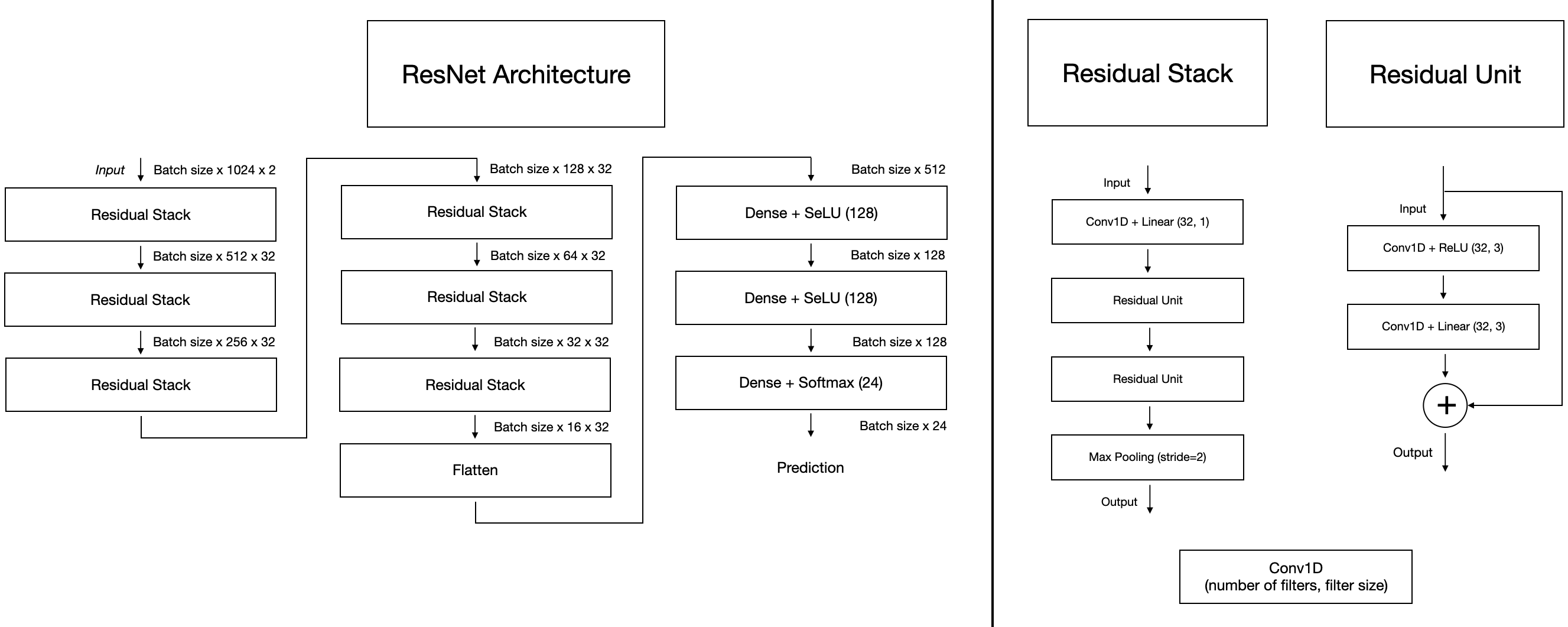}
        \caption{ResNet architecture used in \cite{deepsig}. Each block represents a unit in the network, which may be comprised of several layers and connections as shown on the right of the figure. Dimensions of the tensors on the output of each block are also shown where appropriate.}
        \label{fig::resnet_arch}
    \end{figure*}
    
    While other works have contributed to increased AMC performance, the importance of many design elements for AMC remains unclear and a number of architectural elements have yet to be investigated. Therefore, in this work, we aim to formalize the impact of a variety of architectural changes and model design decisions on AMC performance.  Numerous modifications to architectures from previous works, including our own \cite{asilomar_2020}, and novel combinations of elements applied to AMC are considered. After an initial investigation, we provide a comprehensive ablation study in this work to investigate the performance impact of various architectural modifications.  Additionally, we achieve new state-of-the-art classification performance on the RadioML 2018.01A dataset \cite{deepsig-data}. Using the best performing model, we provide additional analyses that characterize its performance across modulation modes and signal burst duration.

    \section{Related Work}

The area of AMC has been investigated by several research groups.  We provide a summary of results in AMC to provide context and motivation for our contributions to AMC and the corresponding ablation study described in this paper.

Corgan \textit{et al.} \cite{deepsig_original} illustrate that deep convolutional neural networks are able to achieve high classification performance particularly at low SNRs on a dataset comprising 11 different types of modulation.  It was found that CNNs exceeded performance over expertly crafted features.  Comparing results with architectures in \cite{deepsig_original} and \cite{deepsig}, \cite{liu2021self} improved AMC performance utilizing self-supervised contrastive learning.  First, an encoder is pre-trained in a self-supervised manner through creating contrastive pairs with data augmentation.  By creating different views of the input data through augmentation, contrastive loss is used to maximize the cosine similarity between positive pairs (augmented views of the same input).  Once converged, the encoder is frozen (\textit{i.e.}, the weights are set to fixed values) and two fully-connected layers are added following the encoder to form the classifier.  The classifier is trained using supervised learning to predict the 11 different modulation schemes.  Chen \textit{et al.} applied a novel architecture to the same dataset where the input signal is sliced and transformed into a square matrix and apply a residual network to predict the modulation schemes \cite{9580446}.  Other work has investigated empirical and variational mode decomposition to improve few-shot learning for AMC \cite{9935275}. In our work, we utilize a larger, more complex dataset consisting of 24 modulation schemes, as well as  modeling improvements.

Spectrograms and I/Q constellation plots in \cite{asilomar} were found to be effective input features to a traditional CNN achieving nearly equivalent performance as the baseline CNN network in \cite{deepsig} which used raw I/Q signals.

Further, \cite{Peng, ConstellationDiagrams, GraphicConstellations} also used I/Q constellations as an input feature in their machine learning models on a smaller scale of four or eight modulation types. Other features have been used in AMC---\cite{CPark, Teng} utilized statistical features and support vector machines while \cite{Fusion1, fusion2} used fusion methods in CNN classifiers.  Mao \textit{et al.} utilized various constellation diagrams at varying symbol timings alleviating symbol timing synchronization concerns \cite{mao2021attentive}.  A squeeze-and-excitation \cite{hu2018squeeze} inspired architecture was used as an attention mechanism to focus on the most important diagrams.

Although spectrograms and constellation plots have shown promise, they require additional processing overhead and have had comparable performance to raw I/Q signals.  In addition, models that use raw I/Q signals could be more adept at handling varying-length signals than constellation plots because they are not limited by periodicity constraints for short duration signals (\textit{i.e.}, burst transmissions).  Consequently, we utilize raw I/Q signals in our work.

Tridgell, in his dissertation \cite{fgpa}, builds upon these works by investigating these architectures when deployed on resource-limited Field Programmable Gate Arrays (FGPAs).  His work stresses the importance of reducing the number of parameters for modulation classifiers because they are typically deployed in resource-constrained embedded systems.

\begin{figure}[htbp]
    \centering
    \includegraphics[width=\columnwidth]{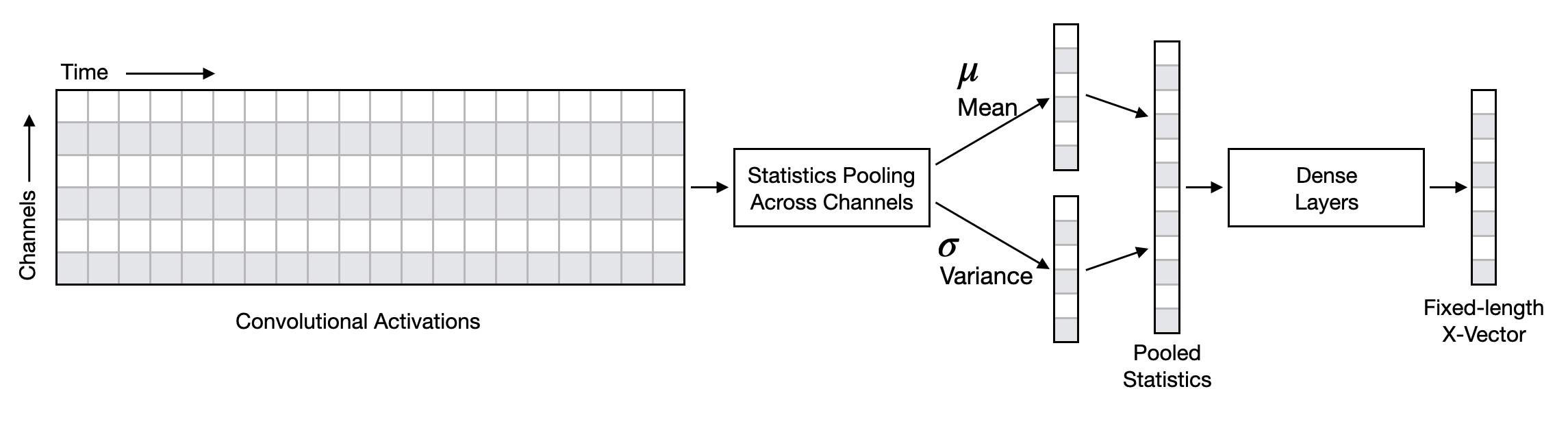}
    \caption{X-Vector architecture overview. The convolutional activations immediately before pooling are shown. These activations are fed into two statistical pooling layers that collapse the activations over time, creating a fixed-length tensor that can be further processed by fully connected dense layers.}
    \label{fig::x_vec_overview}
\end{figure}

\begin{figure*}[htbp]
    \centering
    \includegraphics[width=2\columnwidth]{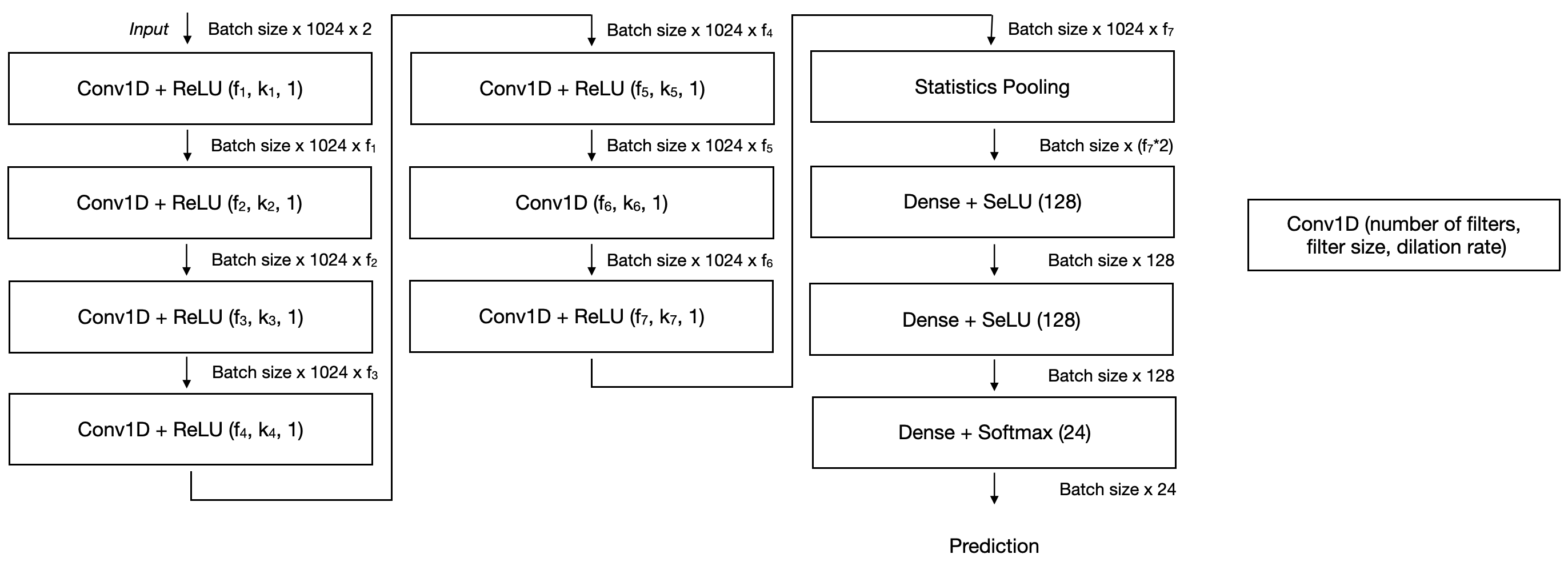}
    \caption{Proposed CNN Architecture in \cite{asilomar_2020}. This is the first work to employ an X-Vector inspired architecture for AMC showing strong performance. This architecture is used as a baseline for the modifications investigated in this paper. The $f$ and $k$ variables shown designate the number of kernels and size of each kernel, respectively, in each layer. These parameters are investigated for optimal sizing in our initial investigation. }
    \label{fig::xvec_architecture}
\end{figure*}

In \cite{deepsig}, Oshea \textit{et al.} created a dataset with 24 different types of modulation, known as RadioML 2018.01A, and achieved high classification performance using convolutional neural networks—specifically using residual connections (see \Cref{fig::resnet_arch}) within the network (ResNet).  A total of 6 residual stacks were used in the architecture.  A residual stack is defined as a series of a convolutional layers, residual units, and a max pooling operation as shown in \Cref{fig::resnet_arch}. The ResNet employed by \cite{deepsig} attained approximately 95\% classification accuracy at high SNR values.

Harper \textit{et al.} proposed the use of X-Vectors \cite{snyder2018x} to increase classification performance using CNNs \cite{asilomar_2020}. X-Vectors are traditionally used in speaker recognition and verification systems making use of aggregate statistics. X-Vectors employ statistical moments, specifically mean and variance, across convolutional filter outputs.  It can be theorized that taking the mean and variance of the embedding layer helps to eliminate signal-specific information, leaving global, modulation-specific characteristics.  \Cref{fig::x_vec_overview} illustrates the X-Vector architecture where statistics are computed over the activations from a convolutional layer producing a fixed-length vector.

Additionally, this architecture maintains a fully-convolutional structure enabling variable size inputs into the network.  Using statistical aggregations allows for this property to be exploited.  When using statistical aggregations, the input to the first dense layer is dependent upon the number of filters in the final convolutional layer.  The number of filters is a hyperparameter, independent of the length in time of the input signal into the neural network. 

Without the statistical aggregations, the input signals into a traditional CNN or ResNet would need to be resampled, cropped or padded to a fixed-length in time such that there is not a size mismatch with the final convolutional output and the first dense layer.  While the dataset used in this work has uniformly sized signals in terms of duration, ($1024\times2$), this is an architectural advantage in our deployment as received signals may vary in duration.  Instead of modifying the inputs to the network via sampling, cropping, padding, etc., the X-Vector architecture can directly operate with variable-length inputs without modifications to the network or input signal.

\Cref{fig::xvec_architecture} outlines the employed X-Vector architecture in \cite{asilomar_2020} where $F = [f_1, f_2, ..., f_7] = 64$ and $K = [k_1, k_2, ..., k_7] = 3$.  Mean and variance pooling are performed on the final convolutional outputs, concatenated, and fed through a series of dense layers creating the fixed-length X-Vector.  A maximum of 98\% accuracy was achieved at high SNR levels.

\begin{figure}[htbp]
\centering
\includegraphics[width=\columnwidth]{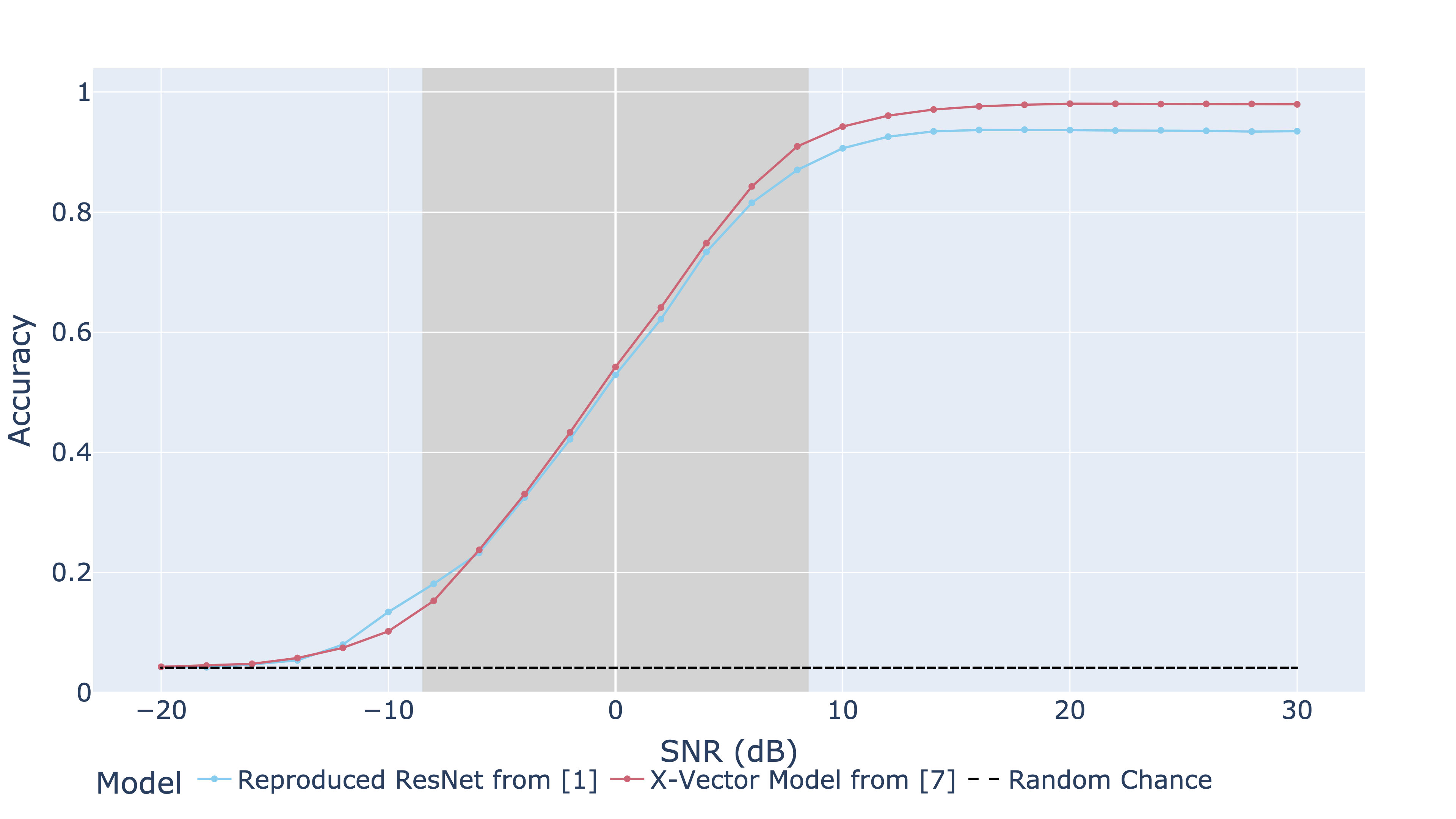}
\caption{Accuracy comparison of the reproduced ResNet in \cite{deepsig} and the X-Vector inspired model from \cite{asilomar_2020} over varying SNRs. This accuracy comparison shows the superior performance of the X-Vector architecture, especially at higher SNRs, and supports using this architecture as a baseline for the improvements investigated in this paper. }
\label{fig::asil_2020_curve}
\end{figure}
The work of \cite{asilomar_2020} replicated the ResNet architecture from \cite{deepsig} and compared the results with the X-Vector architectures as seen in \Cref{fig::asil_2020_curve}.  Harper \textit{et al.} \cite{asilomar_2020} were able to reproduce this architecture achieving a maximum of 93.7\% accuracy.  The authors attribute the difference in performance to differences in the train and test set separation they used since these parameters were unavailable.
As expected, the classifiers perform with a higher accuracy as the SNR value increases.  In signals with a low SNR value, noise becomes more dominant and the signal is harder to distinguish.  In modern software-defined radio applications, a high SNR value is not always a given. However, there is still significant improvement compared to random chance, even at low SNR values.  Moreover, in systems where the modulation type must be classified quickly, this could become crucially important as fewer demodulation schemes would need to be applied in a trial and error manner to discover the correct scheme.

One challenge of AMC is that performance is desired to work well across a large range of SNRs.  For instance, \Cref{fig::asil_2020_curve} illustrates modulation classification performance plateaued in peak performance beyond $+8$dB SNR and approached chance classification performance below $-8$dB SNR on the RadioML 2018.01A dataset. This range is denoted by the shaded region.  Harper \textit{et al.} investigated methods to improve classification performance in this range by employing an SNR regression model to aid separate modulation classifiers (MCs). 
While other works have trained models to be as resilient as possible under varying SNR conditions, Harper \textit{et al.} employed SNR-specific MCs \cite{harper2021snr}.

\begin{table}[htbp]
\caption{SNR groupings for Training SNR-Specific Classifiers and Demultiplexed Classification Ranges for Each Predicted SNR.}
\centering
\begin{tabular}{cc} 
\multicolumn{1}{l}{Training Range (dB)} & \multicolumn{1}{l}{Demultiplexed Classification Range (dB)} \\ \hline
{[}-20, -8{]}                           & ($-\infty$, -8)                                   \\ 
{[}-8, -4{]}                            & {[}-8, -4)                                   \\ 
{[}-4, 0{]}                             & {[}-4, 0)                                    \\ 
{[}0, 4{]}                              & {[}0, 4)                                     \\ 
{[}4, 8{]}                              & {[}4, 8)                                     \\ 
{[}8, 30{]}                             & {[}8, $\infty$)     \\ \hline 
\end{tabular}\label{table::db_groupings}
\end{table}

\begin{figure}[htbp]
\centering
\includegraphics[width=\columnwidth]{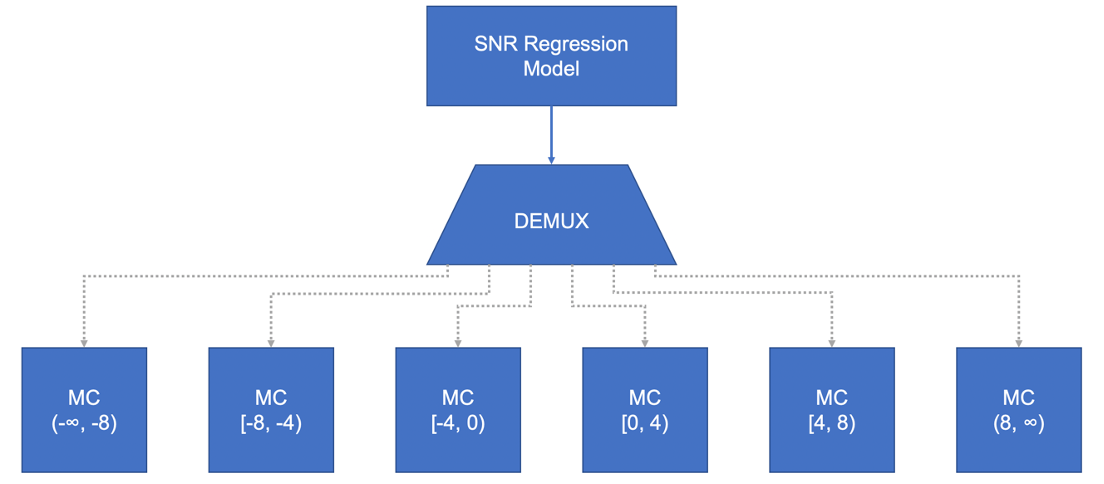}
\caption{The architecture using SNR regression and SNR-specific classifiers from \cite{harper2021snr}. Each MC block shown employs the same architecture as the baseline from \cite{asilomar_2020}, but specifically trained to perform AMC within a more narrow range of SNRs (denoted as dB ranges in each block). }
\label{fig::demux}
\end{figure}

Six MCs were created by discretizing the SNR range to ameliorate performance between $-8$dB to $+8$dB SNR (see \Cref{fig::demux}).  These groupings were chosen in order to provide sufficient training data to avoid overfitting the MCs and provide enough resolution so that combining MCs provided more value than a single classifier.

By first predicting the SNR of the received signal with a regression model, an SNR-specific MC that was trained on signals with the predicted SNR is applied to make the final prediction. Although the SNR values in the dataset are discrete, SNR is measured on a continuous scale in a deployment scenario and can vary over time.  As a result, regression is used over classification to model SNR.  Using this approach, different classifiers can tune their feature processing for differing SNR ranges.  Each MC in this approach uses the same architecture as that proposed in \cite{asilomar_2020}; however, each MC is trained with signals within each MC's SNR training range (see \Cref{table::db_groupings}).

\begin{figure}[htbp]
\centering
\includegraphics[width=\columnwidth]{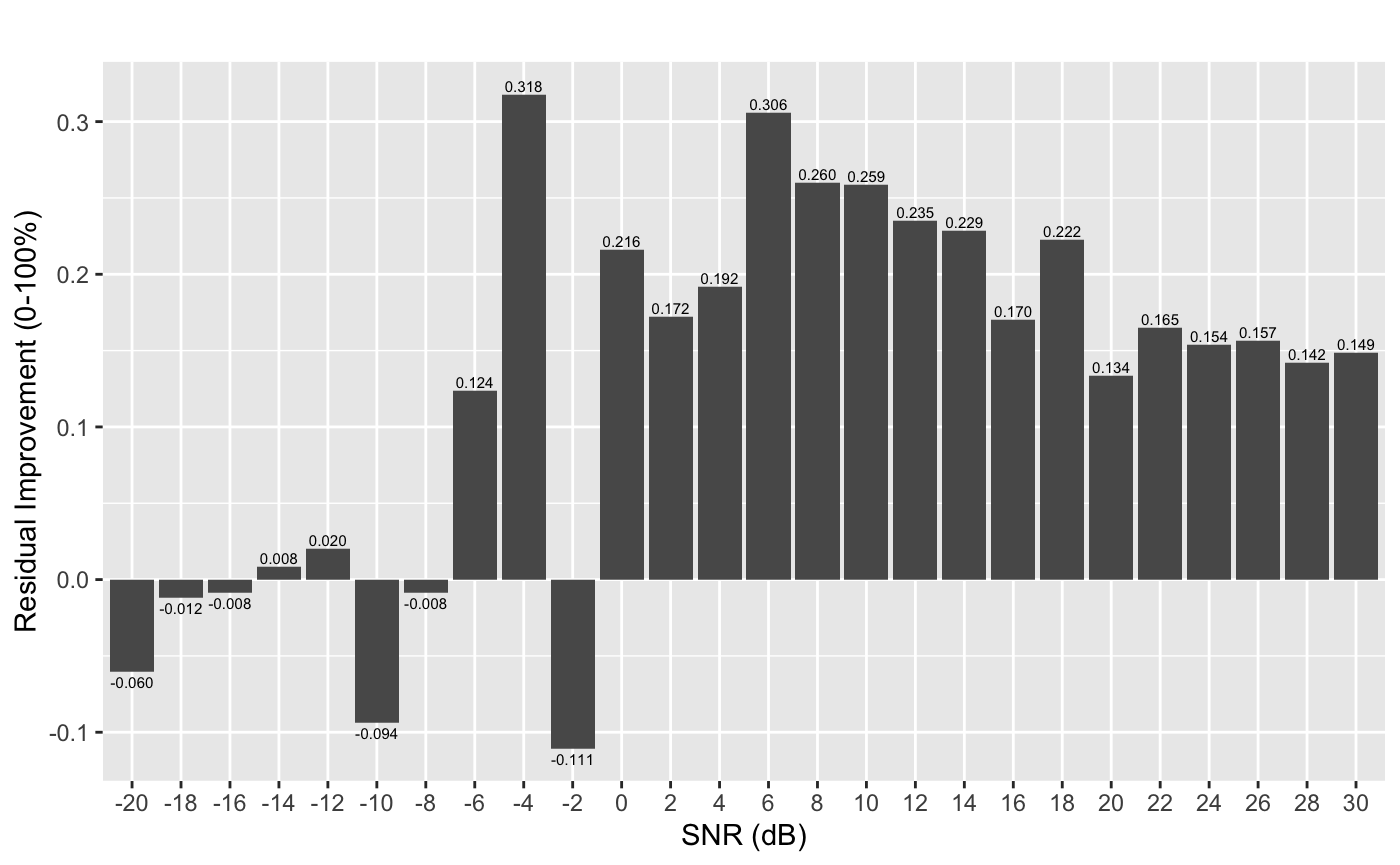}
\caption{Summary of residual improvement in accuracy over \cite{asilomar_2020} that was first published in \cite{harper2021snr}. This work showed how the baseline architecture could be tuned to specific SNR ranges. Positive improvement is observed for most SNR ranges. }
\label{fig::pct_improvement}
\end{figure}

Highlighting improvements across varying SNR values, \Cref{fig::pct_improvement} shows the overall performance improvement (in percentage accuracy) using the SNR-assisted architecture compared to the baseline classification architecture described in \cite{asilomar_2020}.  While a slight decrease in performance was observed for $-8$dB and a larger decrease for $-2$dB, improvement is shown under most SNR conditions---particularly in the target range of $-8$dB to $+8$dB. 
A possible explanation for the decrease in performance at particular SNRs is that the optimization for a particular MC helped overall performance for a grouping at the expense of a single value in the group.  That is, the MC for $[-4, 0)$ boosted the overall performance by performing well at $-4$ and $0$dB at the expense of $-2$dB.  Due to the large size of the testing set, these small percentage gains are impactful because thousands more classifications are correct.  All results are statistically significant based on a McNemar's test \cite{mcnemar1947note}, therefore achieving new state-of-the-art performance at the time.

Soltani \textit{et al.}\cite{soltani2019spectrum} found SNR regions of $[-10, -2]$dB, $[0, 8]$dB, and $[10, 30]$dB having similar classification patterns.  Instead of predicting exact modulation variants, the authors group commonly confused variants into a more generic, coarse-grained label. This grouping increases performance of AMC by combining modulation variants that are commonly confused. However, it also decreases the sensitivity of the model to the numerous possible variants.

Cai \textit{et al.} utilized a transformer based architecture to aid performance at low SNR levels with relatively few training parameters (approximately 265,0000 parameters) \cite{9779340}.  A multi-scale network along with center loss \cite{wen2016discriminative} was used in \cite{9463441}.  It was found that larger kernel sizes improved AMC performance.  We further explore kernel size performance impacts in this work.  Zhang \textit{et al.} proposed a high-order attention mechanism using the covariance matrix achieving a maximum accuracy of 95.49\% \cite{9986040}.

Although many discussed works use the same RadioML 2018.01A dataset, there is a lack of a uniform dataset split to establish a benchmark for papers to report performance. In an effort to make AMC work more reproducible and comparable across publications, we have made our dataset split and accompanying code available on GitHub.\footnote{\url{https://github.com/caharper/Automatic-Modulation-Classification-with-Deep-Neural-Networks}{}}

While numerous works have investigated architectural improvements, we aim to improve upon these works by introducing additional modifications as well as a comprehensive ablation study that illustrates the improvement of each modification. With the new modifications, we achieve new state-of-the-art AMC performance.  

\section{Dataset}\label{sec::dataset}
To evaluate different machine learning architectures, we use the RadioML 2018.01A dataset that is comprised of 24 different modulation types \cite{deepsig, deepsig-data}.  
Due to the complexity and variety of modulation schemes in the dataset, it is fairly representative of typically encountered modulation schemes. Moreover, this variety increases the likelihood that AMC models will generalize to more exotic or non-existing modulation schemes in the training data that are derived from these traditional variants. 

There are a total of 2.56 million labeled signals, $S(T)$, each consisting of 1024 time domain digitized intermediate frequency (IF) samples of in-phase ($I$) and quadrature ($Q$) signal components where $S(T)=I(T)+jQ(T)$.  The data was collected at a 900MHz IF with an assumed sampling rate of 1MS/sec such that each 1024 time domain digitized I/Q sample is 1.024 ms \cite{o2016convolutional}.  The 24 modulation types and the representative groups that we chose for each are listed as follows: 

\begin{itemize}
    \item{\textbf{Amplitude}: OOK, 4ASK, 8ASK, AM-SSB-SC, AM-SSB-WC, AM-DSB-WC, and AM-DSB-SC}
    \item{\textbf{Phase}: BPSK, QPSK, 8PSK, 16PSK, 32PSK, and OQPSK}
    \item{\textbf{Amplitude and Phase}: 16APSK, 32APSK, 64APSK, 128APSK, 16QAM, 32QAM, 64QAM, 128QAM, and 256QAM}
    \item{\textbf{Frequency}: FM and GMSK}
\end{itemize}
 
Each modulation type includes a total of $106,496$ observations ranging from $-20$dB to $+30$dB SNR in $2$dB steps for a total of 26 different SNR values.  SNR is assumed to be consistent over the same window length as the I/Q sample window. 
For evaluation, we divided the dataset into 1 million different training observations and 1.5 million testing observations under a random shuffle split, stratified across modulation type and SNR.  Because of this balance, the expected performance for a random chance classifier is 1/24 or 4.2\%.  With varying SNR levels across the dataset, it is expected that the classifier would perform with a higher degree of accuracy as the SNR value is increased.  For consistency, each model investigated in this work was trained and evaluated on the same train and test set splits. 

\section{Initial Investigation}\label{sec::init_investigation}
In this work, we use the architecture described in \cite{asilomar_2020} as the baseline architecture.  We note that \cite{harper2021snr} improved upon the baseline; however, each individual MC used the baseline architecture except trained on specific SNR ranges.
Therefore, the base architectural elements were similar to \cite{asilomar_2020}, but separated for different SNRs. In this work, our focus is to improve upon the employed CNN architecture for an individual MC rather than the use of several MCs.  Therefore, we use the architecture from \cite{asilomar_2020} as our baseline.

Before exploring an ablation study, we make a few notable changes from the baseline architecture in an effort to increase AMC performance.  This initial exploration is for clarity as it reserves the ablation study that follows from requiring an inordinate number of models. It also introduces the general training procedures that assist and orient the reader in following the ablation study---the ablation study mirrors these procedures. We first provide an initial investigation exploring these notable changes. 

We train each model using the Adam optimizer \cite{kingma2014adam} with an initial learning rate \textit{lr} = 0.0001, a decay factor of 0.1 if the validation loss does not decrease for 12 epochs, and a minimum learning rate of 1e-7.  If the validation loss does not decrease after 20 epochs, training is terminated and the models are deemed converged.  For all experiments, mini-batches of size 32 are used. As has been established in most programming packages for neural networks, we refer to fully connected neural network layers as \textit{dense} layers, which are typically followed by an activation function. 

\subsection{Architectural Changes}
A common property of neural networks is using fewer but larger kernels in the early layers of the network, and an increase of smaller kernels are used in the later layers than the baseline architecture. This is commonly referred to as the information distillation pipeline \cite{chollet2021deep}. By utilizing a smaller number of large kernels in early layers, we are able to increase the temporal context of the convolutional features without dramatically increasing the number of trainable parameters.  Numerous, but smaller kernels are used in later convolutional layers to create more abstract features.  Configuring the network in this manner is especially popular in image classification where later layers represent more abstract, class-specific features. 

We investigate this modification in three stages, using the baseline architecture described in \Cref{fig::xvec_architecture} \cite{asilomar_2020}. We denote number of filters in the network and the filter sizes as $F=[f_1,f_2,...,f_7]$ and $K=[k_1,k_2,...k_7]$ in \Cref{fig::xvec_architecture}. The baseline architecture used $f = 64$ (for all layers) and $k = 3$ (consistent kernel size for all layers).  Our first modification to the baseline architecture is $F = [32, 48, 64, 72, 84, 96, 108]$, but keeping $k = 3$ for all layers.  Second, we use the baseline architecture, but change the size of filters in the network where $f = 64$ (same as baseline) and $K = [7, 5, 7, 5, 3, 3, 3]$.  Third, we make both modifications and compare the result to the baseline model where $F = [32, 48, 64, 72, 84, 96, 108]$ and $K = [7, 5, 7, 5, 3, 3, 3]$. These modifications are not exhaustive searches; rather, these modifications are meant to guide future changes to the network by understanding the influence of filter quantity and filter size in a limited context.

\begin{table}[htbp]
\caption{Initial investigation performance overview. All architectures employ the baseline with varying numbers of kernels and kernel sizes. }
\centering
\begin{tabular}{cccc}
Notes                                                                           & \# Params & \begin{tabular}[c]{@{}c@{}}Avg. \\ Accuracy\end{tabular} & \begin{tabular}[c]{@{}c@{}}Max\\ Accuracy\end{tabular} \\ \hline
Reproduced ResNet \cite{deepsig}                                                       & 165,144   & 59.2\%                                                   & 93.7\%                                                 \\\hline
X-Vector in \cite{asilomar_2020}                                                            & 110,680   & 61.3\%                                                   & 98.0\%                                                 \\\hline
\begin{tabular}[c]{@{}c@{}}More Filters\\ (Same Filter Sizes)\end{tabular}      & 149,168   & 61.0\%                                                   & 96.1\%                                                 \\\hline
\begin{tabular}[c]{@{}c@{}}Larger Filter Sizes\\ (Same \# Filters)\end{tabular} & 143,960   & 62.6\%                                                   & 98.2\%                                                 \\\hline
Combined                                                                        & 174,000   & 62.9\%                                                   & 98.6\%  \\\hline                                              
\end{tabular}\label{table::inital_max_acc}
\end{table}

\subsection{Initial Investigation Results}
As shown in \Cref{table::inital_max_acc}, increasing the size of the filters in earlier layers increases both average and maximum test accuracy over \cite{asilomar_2020}; but, at the cost of additional parameters.  A possible explanation for the increase in performance is the increase in temporal context due to the larger kernel sizes.  Increasing the number of filters without increasing temporal context decreases performance. This is possibly because it increases the complexity of the model without adding additional signal context.

\begin{figure}[htbp]
    \centering
    \includegraphics[width=\columnwidth]{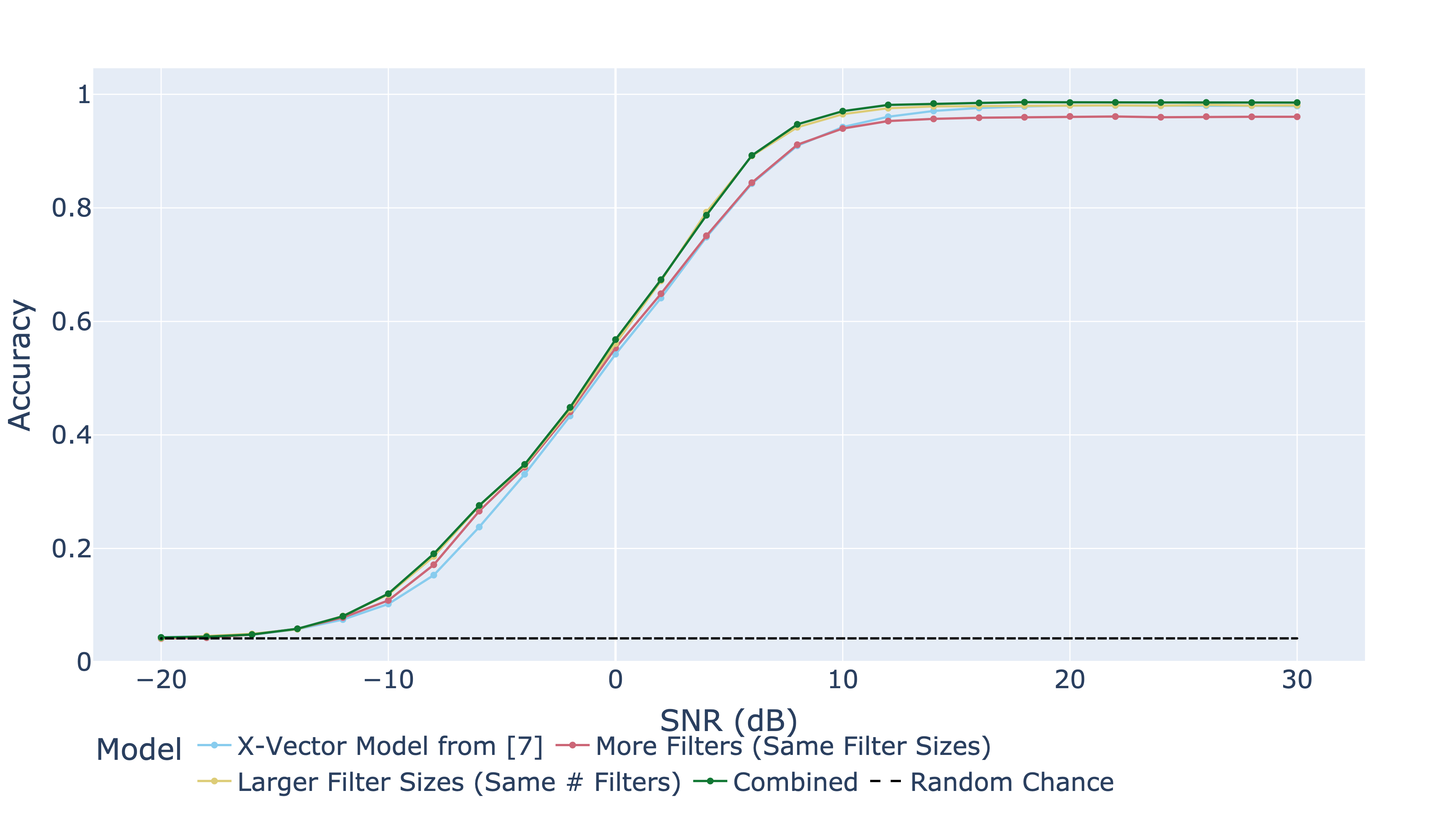}
    \caption{SNR vs. accuracy comparison of the initial investigation using the baseline architecture. Noticeable improvements can be observed across all SNRs. }
    \label{fig::init_invest_snr_acc}
\end{figure}

\Cref{fig::init_invest_snr_acc} illustrates the change in accuracy with varying SNR.  The combined model, utilizing various kernel sizes and numbers of filters, consistently outperforms the other architectures across changing SNR conditions.

Although increasing the number of filters decreases performance alone, combining the approach with larger kernel sizes yields the best performance in our initial investigation.  Increasing the temporal context may have allowed additional filters to better characterize the input signal.

Because increased temporal context improves AMC performance, we are inspired to investigate additional methods such as squeeze-and-excitation blocks and dilated convolutions that can increase global and local context \cite{hu2018squeeze,yu2015multi}.

\section{Ablation Study Architecture Background}
Building upon our findings from our initial investigation, we make additional modifications to the baseline architecture.  For the MCs, we introduce dilated convolutions, squeeze-and-excitation blocks, self-attention, and other architectural changes. We also investigate various kernel sizes and the quantity of kernels employed from the initial investigation.  Our goal is to improve upon existing architectures while investigating the impact of each modification on classification accuracy through an ablation study. In this section, we describe each modification performed.

\subsection{Squeeze-and-Excitation Networks}

\begin{figure}[htbp]
\centering
\includegraphics[width=\columnwidth]{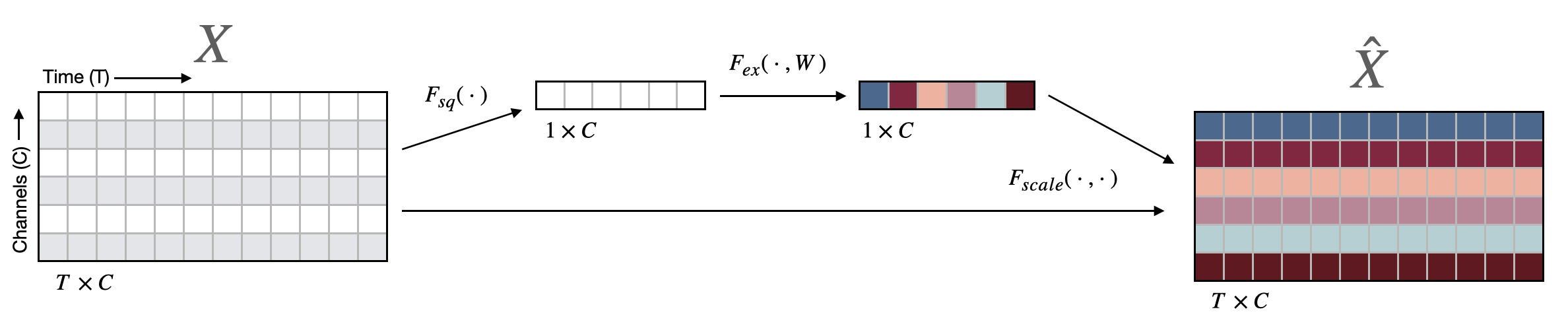}
\caption{Squeeze-and-Excitation block proposed in \cite{hu2018squeeze}. One SE block is shown applied to a single layer convolutional output activation. Two paths are shown, a scaling path and an identity path.  The scaling vector is applied across channels to the identity path of the activations. }
\label{fig::se_block}
\end{figure}

Squeeze-and-Excitation (SE) blocks introduce a channel-wise attention mechanism first proposed in \cite{hu2018squeeze}.  Due to the limited receptive field of each convolutional filter, SE blocks propose a recalibration step based on global statistics across channels (average pooling) to provide global context.  Although initially utilized for image classification tasks \cite{hu2018squeeze, zhang2018shufflenet, tan2019efficientnet}, we argue the use of SE blocks can provide meaningful global context to the convolutional network used for AMC over the time domain.  

\begin{figure*}[t]
    \centering
    \includegraphics[width=1.\textwidth]{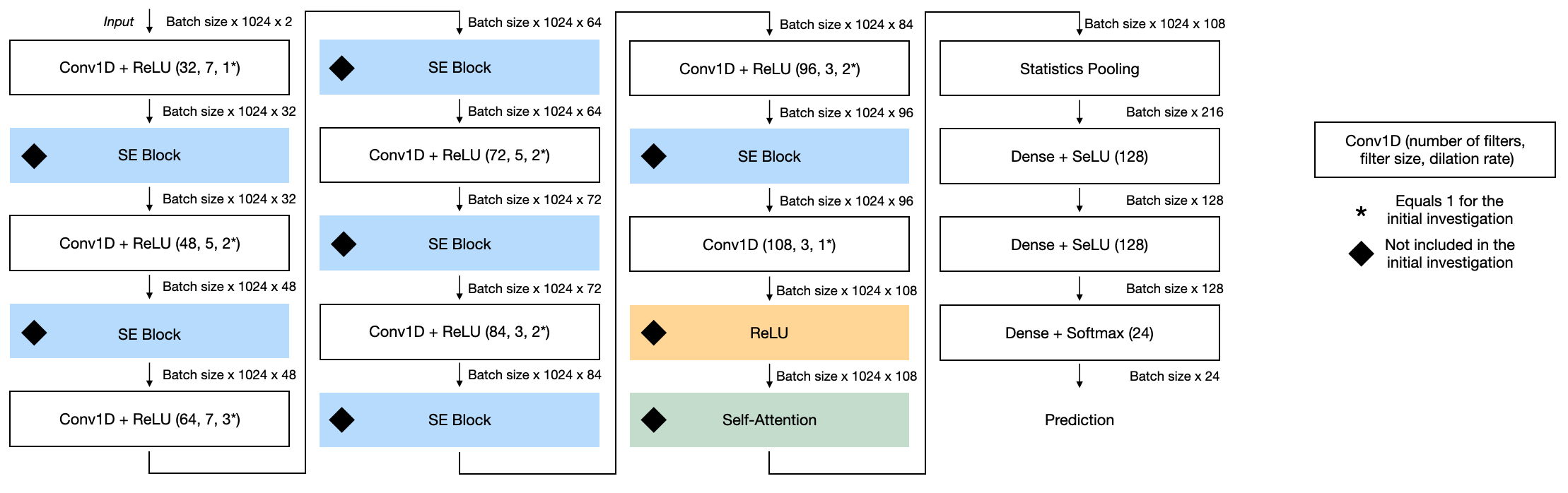}
    \caption{Proposed architecture with modifications including SENets, dilated convolutions, optional ReLU activation before statistics pooling, and self-attention. The output tensor sizes are also shown for each unit in the diagram. An * denotes where the sizes differ from the baseline architecture.  }
    \label{fig::proposed_arch_fig}
\end{figure*}

\Cref{fig::se_block} depicts an SE block.  The squeeze operation is defined as temporal global average pooling across convolutional filters.  For an individual channel, $c$, the squeeze operation is defined as: 
\begin{equation}
    z_c = F_{sq}(x_c) = \frac{1}{T}\sum_{i=1}^{T}x_{i,c}
  \label{eqn::squeeze}
\end{equation}
where $X \in \mathbb{R}^{T \times C} = [x_1, x_2, ..., x_C]$, $Z \in \mathbb{R}^{1 \times C} = [z_1, z_2, ..., z_C]$, $T$ is the number of samples in time, and $C$ is the total number of channels.  To model nonlinear interactions between channel-wise statistics, $Z$ is fed into a series of dense layers followed by nonlinear activation functions:
\begin{equation}
    s = F_{ex}(z, W) = \sigma(g(z, W)) = \sigma(W_2\delta(W_1z))
  \label{eqn::excitation}
\end{equation}
where $\delta$ is the rectified linear (ReLU) activation function, $W_1 \in \mathbb{R}^{\frac{C}{r} \times C}$, $W_2 \in \mathbb{R}^{C \times \frac{C}{r}}$, $r$ is a dimensionality reduction ratio, and $\sigma$ is the sigmoid activation function.  The sigmoid function is chosen as opposed to the softmax function so that multiple channels can be accentuated and are not mutually-exclusive. That is, the normalization term in the softmax can cause dependencies among channels, so the sigmoid activation is preferred.  $W_1$ imposes a bottleneck to improve generalization performance and reduce parameter counts while $W_2$ increases the dimensionality back to the original number of channels for the recalibration operation. In our work, we use $r=2$ for all SE blocks to ensure a reasonable number of trainable parameters without over-squashing the embedding size.

The final operation in the SE block, scaling or recalibration, is obtained by scaling the the input $X$ by $s$:

\begin{equation}
    \hat{x_c} = F_{scale}(x_c, s_c) = s_c x_c
  \label{eqn::scaling}
\end{equation}
where $\hat{X} \in \mathbb{R}^{T \times C} = [\hat{x_1}, \hat{x_2}, ..., \hat{x_C}]$. 


\subsection{Dilated Convolutions}

\begin{figure}[htbp]
\centering
\includegraphics[width=\columnwidth]{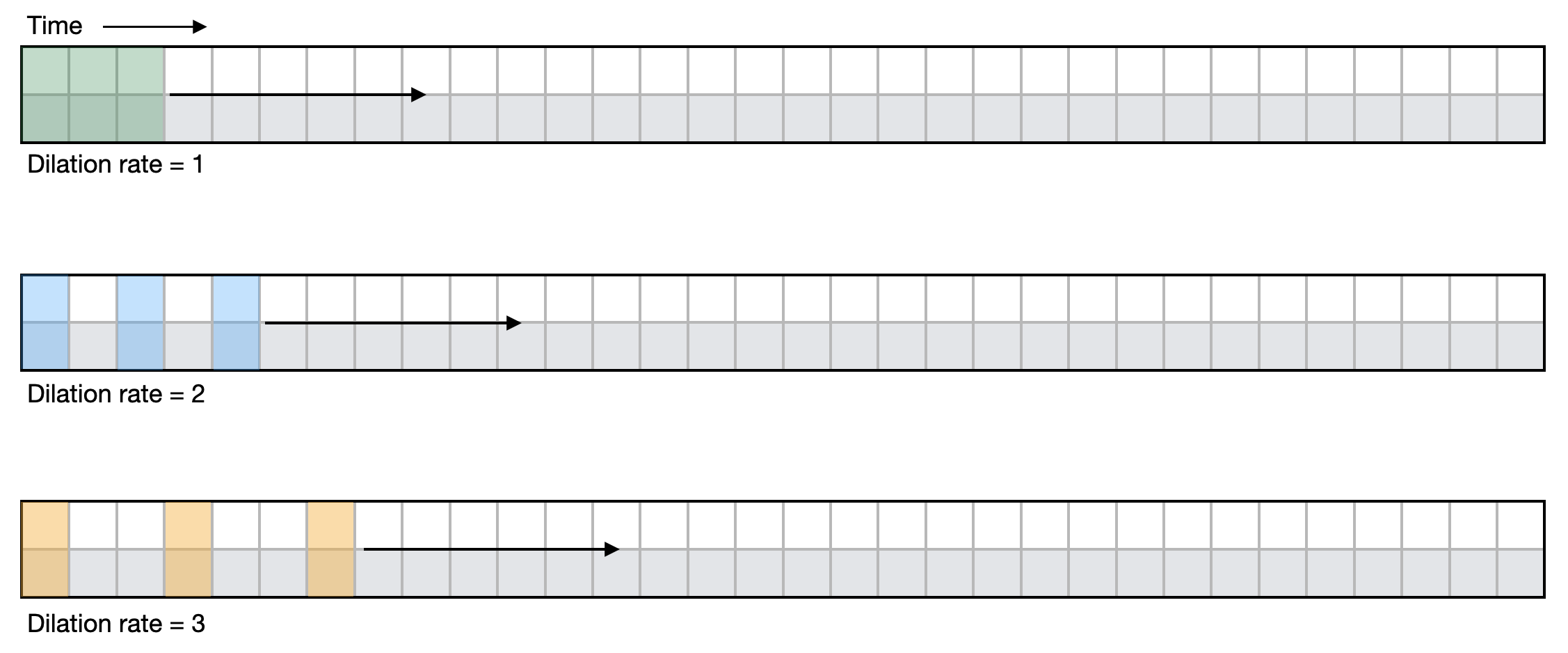}
\caption{Dilated convolutions diagram. The top shows a traditional kernel applied to sequential time series points. The middle and bottom diagram illustrate dilation rates of two and three, respectively. These dilations serve to increase the receptive field of the filter without increasing the number of trainable variables in the kernel. }
\label{fig::dilated_convs}
\end{figure}

Proposed in \cite{yu2015multi}, \Cref{fig::dilated_convs} depicts dilated convolutions where the convolutional kernels are denoted by the colored components.  In a traditional convolution, the dilation rate is equal to 1. Dilated convolutions build temporal context by increasing the receptive field of the convolutional kernels without increasing parameter counts as the number of entries in the kernel remains the same.

Dilated convolutions also do not downsample the signals like strided convolutions. Instead, the output of a dilated convolution can be the exact size of the input after properly handling edge effects at the beginning and end of the signal.

\subsection{Final Convolutional Activation}
We also investigate the impact of using an activation function (ReLU) after the last convolutional layer, just before statistics pooling.  Because ReLU transforms the input sequence to be non-negative, the distribution characterized by the pooling statistics may become skewed.  In \cite{asilomar_2020} and \cite{harper2021snr}, no activation was applied after the final convolutional layer as shown in \Cref{fig::xvec_architecture}.
We investigate if this transformation impacts classification performance.

\subsection{Self-Attention}

Self-attention allows the convolutional outputs to interact with one another enabling the network to learn to focus on important outputs.  Self-attention before statistics pooling essentially creates a weighted summation over the convolutional outputs weighting their importance similarly to \cite{okabe2018attentive, safari20_interspeech, sammit2022prosody}. 

We use the attention mechanism described by Vaswani \textit{et al.} in \cite{vaswani2017attention} where each output element is a weighted sum of the linearly transformed input where the dimensionality of $K$ is $d_k$ as seen in \Cref{eqn::attn}.  
\begin{equation}
    Attention(Q, K, V) = softmax\left(\frac{QK^T}{|\sqrt{d_k}|}\right)V
  \label{eqn::attn}
\end{equation}
In the case of self-attention, $Q$, $K$, and $V$ are equal.  A scaling factor of $\frac{1}{|\sqrt{d_k}|}$ is applied to counteract vanishing gradients in the softmax output when $d_k$ is large.

\section{Ablation Study Architecture}

    Applying the specified modifications to the architecture in \cite{asilomar_2020},  \Cref{fig::proposed_arch_fig} illustrates the proposed architecture with every modification included in the graphic. Each colored block represents an optional change to the architecture that will be investigated in the ablation study.  That is, each combination of network modifications are analyzed to aid understanding of each modification's impact on the network.
    
    Each convolutional layer has the following parameters: number of filters, kernel size, and dilation rate.  The asterisk next to each dilation rate represents the changing of dilation rates in the ablation study.  If dilated convolutions are used, then the dilation rate value in the graphic is used.  If dilated convolutions are not used, each dilation rate is set to 1.  That is, a traditional convolution is applied. All convolutions use a stride of 1, and the same training procedure from the initial investigation is used.

   \begin{table*}[htbp]
        \caption{Ablation study performance overview.}
            \resizebox{\textwidth}{!}{\begin{tabular}{ccccccccc}
        Model Name & Notes                     & SENet & \begin{tabular}[c]{@{}c@{}}Dilated\\ Convolutions\end{tabular} & \begin{tabular}[c]{@{}c@{}}Final \\ Activation\end{tabular} & Attention & \# Params & \begin{tabular}[c]{@{}c@{}}Avg. \\ Accuracy\end{tabular} & \begin{tabular}[c]{@{}c@{}}Max\\  Accuracy\end{tabular} \\ \hline
        --- & Reproduced ResNet \cite{deepsig} & --- & --- & --- & --- & 165,144 & 59.2\% & 93.7\% \\ \hline
        --- & X-Vector in \cite{asilomar_2020} & --- & --- & --- & --- & 110,680 & 61.3\% & 98.0\% \\ \hline
        0000 & \begin{tabular}[c]{@{}c@{}}Best performing model from\\  the initial investigation\end{tabular} & --- & --- & --- & --- & 174,000 & 62.9\% & 98.6\% \\ \hline
        0001 & & --- & --- & --- & \ding{51} & 221,088 & 62.3\% & 97.6\% \\ \hline
        0010 & & --- & --- & \ding{51} & --- & 174,000 & 62.8\% & 98.6\% \\ \hline
        0011 & & --- & --- & \ding{51} & \ding{51} & 221,088 & 62.3\% & 97.5\% \\ \hline
        0100 & & --- & \ding{51} & --- & --- & 174,000 & 63.2\% & 98.9\% \\ \hline 
        0101 & & --- & \ding{51} & --- & \ding{51} & 221,088 & 63.1\% & 97.9\% \\ \hline
        0110 & & --- & \ding{51} & \ding{51} & --- & 174,000 & 63.2\% & 98.9\% \\ \hline
        0111 & & --- & \ding{51} & \ding{51} & \ding{51} & 221,088 & 63.0\% & 98.0\% \\ \hline
        1000 & & \ding{51} & --- & --- & --- & 202,880 & 62.9\% & 98.5\% \\ \hline
        1001 & & \ding{51} & --- & --- & \ding{51} & 249,968 & 62.6\% & 98.2\% \\ \hline
        1010 & & \ding{51} & --- & \ding{51} & --- & 202,880 & 62.6\% & 98.3\% \\ \hline
        1011 & & \ding{51} & --- & \ding{51} & \ding{51} & 249,968 & 62.8\% & 98.1\% \\ \hline
        1100 & & \ding{51} & \ding{51} & --- & --- & 202,880 & 62.8\% & 98.2\% \\ \hline
        1101 & & \ding{51} & \ding{51} & --- & \ding{51} & 249,968 & 63.0\% & 97.7\% \\ \hline
        1110 & Overall best performing model & \ding{51} & \ding{51} & \ding{51} & --- & 202,880 & 63.7\% & 98.9\% \\ \hline
        1111 & & \ding{51} & \ding{51} & \ding{51} & \ding{51} & 249,968 & 63.0\% & 97.8\% \\ \hline   
    \end{tabular}}
    \label{table::abl_overall_performance}
    \end{table*}

\section{Evaluation Metrics}
    We present several evaluation metrics to compare the different architectures considered in the ablation study.  In this section, we will discuss each evaluation technique used in the results section.
  
    Due to the varying levels of SNRs in the employed dataset, we plot classification accuracy over each true SNR value.  This allows for a visualization of the tradeoff in performance as noise becomes more or less dominant in the received signals.  Additionally, we report average accuracy and maximum accuracy across the entire test set for each model.  While we note that average accuracy is not indicative of the model's performance, as accuracy is highly correlated to the SNR of the input signal, we share this result to give other researchers the ability to reproduce and compare works.

    As discussed in \cite{fgpa}, AMC is often implemented on resource-constrained devices.  In these systems, using larger models in terms of parameter counts may not be feasible.  We report the number of parameters for each model in the ablation study to examine the tradeoff in AMC performance and model size.
    
    Additional analyses are also carried out. However, due to the large number of models investigated in this study, we will select the best performing model from the ablation study for brevity and analyze the performance of this model in greater detail. For example, confusion matrices for the best performing model from the ablation study are provided to show common misclassifications for each modulation type.  Additionally, there exist several use-cases where relatively short signal bursts are received.  For example, a wide-band scanning receiver may only detect a short signal burst. Therefore, signal duration in the time domain versus AMC performance is investigated to determine the robustness of the best performing model when short signal bursts are received.  


\section{Ablation Results}

    \subsection{Overall Performance}
    
    \Cref{table::abl_overall_performance} lists the maximum and average accuracy performance for each model in the ablation study. A binary naming convention is used to indicate the various methods used for each architecture.  Similarly to the result found in \Cref{sec::init_investigation},
    increasing the temporal context typically results in increased performance.  Models that incorporate dilated convolutions tended to have higher average accuracies than models without dilated convolutions.  
    
    The best performing model, in terms of average accuracy across all SNR conditions included SE blocks, dilated convolutions, and a ReLU activation prior to statistics pooling (model 1110) with an average accuracy of approximately 63.7\%.  This model also achieved the highest maximum accuracy of about 98.9\% at a 22dB level.
    
    \begin{figure*}[htbp]
    \centering
    \includegraphics[width=1.\textwidth]{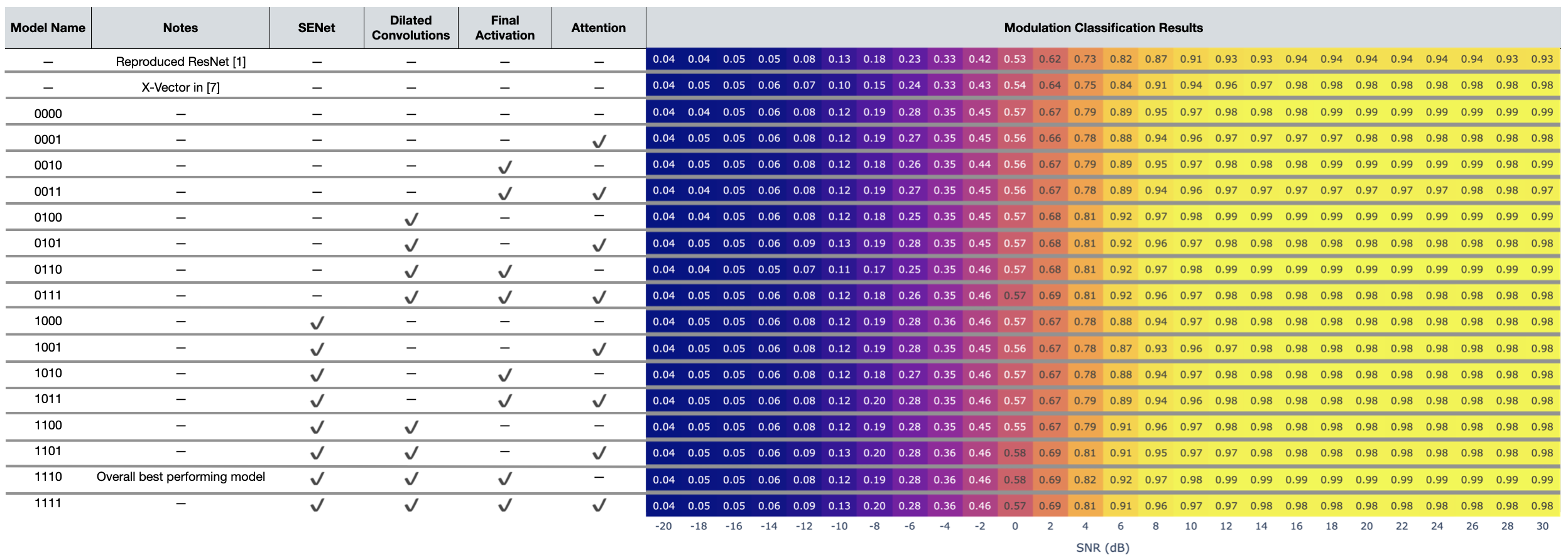}
    \caption{Ablation study results in terms of classification accuracy across SNR ranges. The best performing model is in the second to last row and displays strong performance across SNR values. }
    \label{fig::acc_ablation}
    \end{figure*}
    
    SE blocks did not increase performance compared to model 0000 with the exception of models 1110 and 1111.  However, SE blocks were incorporated in the best performing model, 1110.  Self-attention was not found to aid classification performance in general with the proposed architecture. Self-attention introduces a large number of trainable parameters possibly forming a complex loss space. 
    
    \begin{table}[htbp]
    \caption{Individual network modification performance overview. Entries are repeated from \Cref{table::abl_overall_performance} for clarity. }
    \resizebox{\columnwidth}{!}{
    \begin{tabular}{ccccccccc}
    Model Name & Notes                     & SENet & \begin{tabular}[c]{@{}c@{}}Dilated\\ Convolutions\end{tabular} & \begin{tabular}[c]{@{}c@{}}Final \\ Activation\end{tabular} & Attention & \# Params & \begin{tabular}[c]{@{}c@{}}Avg. \\ Accuracy\end{tabular} & \begin{tabular}[c]{@{}c@{}}Max\\  Accuracy\end{tabular} \\ \hline
    --- & X-Vector in \cite{asilomar_2020} & --- & --- & --- & --- & 110,680 & 61.3\% & 98.0\% \\ \hline
    0000 & & --- & --- & --- & --- & 174,000 & 62.9\% & 98.6\% \\ \hline
    0001 & & --- & --- & --- & \ding{51} & 221,088 & 62.3\% & 97.6\% \\ \hline
    0010 & & --- & --- & \ding{51} & --- & 174,000 & 62.8\% & 98.6\% \\ \hline
    0100 & & --- & \ding{51} & --- & --- & 174,000 & 63.2\% & 98.9\% \\ \hline 
    1000 & & \ding{51} & --- & --- & --- & 202,880 & 62.9\% & 98.5\% \\ \hline
    1110 & Best performer & \ding{51} & \ding{51} & \ding{51} & --- & 202,880 & 63.7\% & 98.9\% \\ \hline
    \end{tabular}\label{table::subset_abl_overall_performance}}
    \end{table}
    
    \Cref{table::subset_abl_overall_performance} lists the performances of single modification (from baseline) architectures. Each component of the ablation study, with the exception of dilated convolutions, decreased performance when applied individually.  When combined, however, the best performing model was found. Therefore, we conclude that each component could possibly aid the optimization of each other---and, in general, dilated convolutions tend to have the most dramatic performance increases. 

\subsection{Accuracy Over Varying SNR}
    \Cref{fig::acc_ablation} summarizes the ablation study in terms of classification accuracy over varying SNR levels.  We add this figure for completeness and reproducibility for other researchers. The accuracy within each SNR band is shown along with the modifications used, similar to \Cref{table::abl_overall_performance}. The coloring in the figure denotes the accuracy in each SNR band. Performance follows a trend similar to that of a sigmoid function, where the rate at which peak classification accuracy is achieved is the most distinguishing feature between the different models.  With the improved architectures, a maximum of 99\% accuracy is achieved at high SNR levels (starting around 12dB SNR). 
    
    While the proposed changes to the architectures generally improve performance at higher SNR levels, the largest improvements occur between $-12$dB and $12$dB compared to the baseline model in \cite{asilomar_2020}. For example, at $4$dB, the performance increases from 75\% up to 82\%. Incorporating these modifications to the network may prove to be critical in real-world situations where noisy signals are likely to be obtained.  Improving AMC performance at lower SNR ranges ($<-12$dB) is  still an open research topic, with accuracies near chance level. 
    
    One observation is the best performing model can vary with SNR.  In systems that have available memory and processing power, an approach similar to \cite{harper2021snr} may be used to utilize several models and intelligently chose predictions based on estimated SNR conditions. That is, if the SNR of the signal of interest is known, a model can be tuned to increase performance slightly, as shown in \cite{harper2021snr}. Using the results presented here, researchers could also choose the architecture differences that perform best for a given SNR range (although performance differences are subtle).

\subsection{Parameter Count Tradeoff}

    \begin{figure}[htbp]
    \centering
    \includegraphics[width=1.\columnwidth]{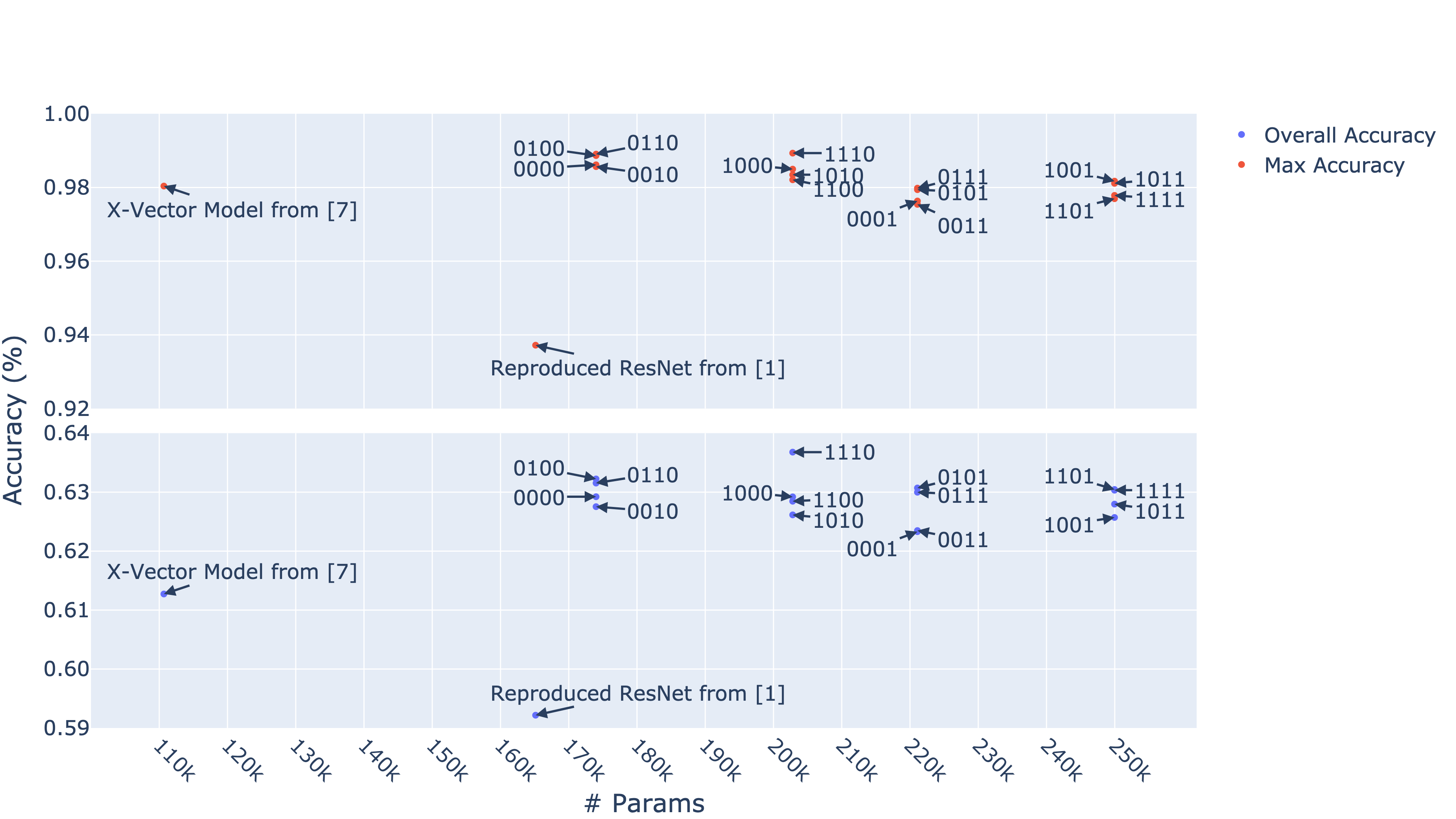}
    \caption{Ablation study parameter count tradeoff. The x-axis shows the number of trainable variables in each model and the y-axis shows max or average accuracy. The callout for each point denotes the model name as shown in \Cref{table::abl_overall_performance}. }
    \label{fig::param_tradeoff}
    \end{figure}
    
    An overview of each model's complexity and overall performance across the entire testing set is shown in \Cref{table::abl_overall_performance}. This information is also shown graphically in \Cref{fig::param_tradeoff} for the maximum accuracy over SNR and the average accuracy across all SNRs. Whether looking at the maximum or the average measures of performance, the conclusions are similar. The previously described binary model name also appears in the figure. We found a slight correlation between the number of model parameters and overall model performance; however, with the architectures explored, there was a general parameter count where performance peaked.  Models with parameter counts between approximately 170k to 205k generally performed better than smaller and larger models.  We note that the models with more than 205k parameters included self-attention which was found to decrease model performance with the proposed architectures. This implies that one possible reason self-attention did not perform as well as other modifications is because of the increase in parameters, resulting in a more difficult loss space from which to optimize.   
    
\section{Best Performing Model Investigation}

    Due to the large volume of models, we focus upon the best performing model, (model 1110), for the remainder of this work. As previously mentioned, this model employs all modifications except self-attention.

\subsection{Top-K Accuracy}

\begin{figure*}[!t]
    \centering
    \subfloat[]{\includegraphics[width=.33\textwidth]{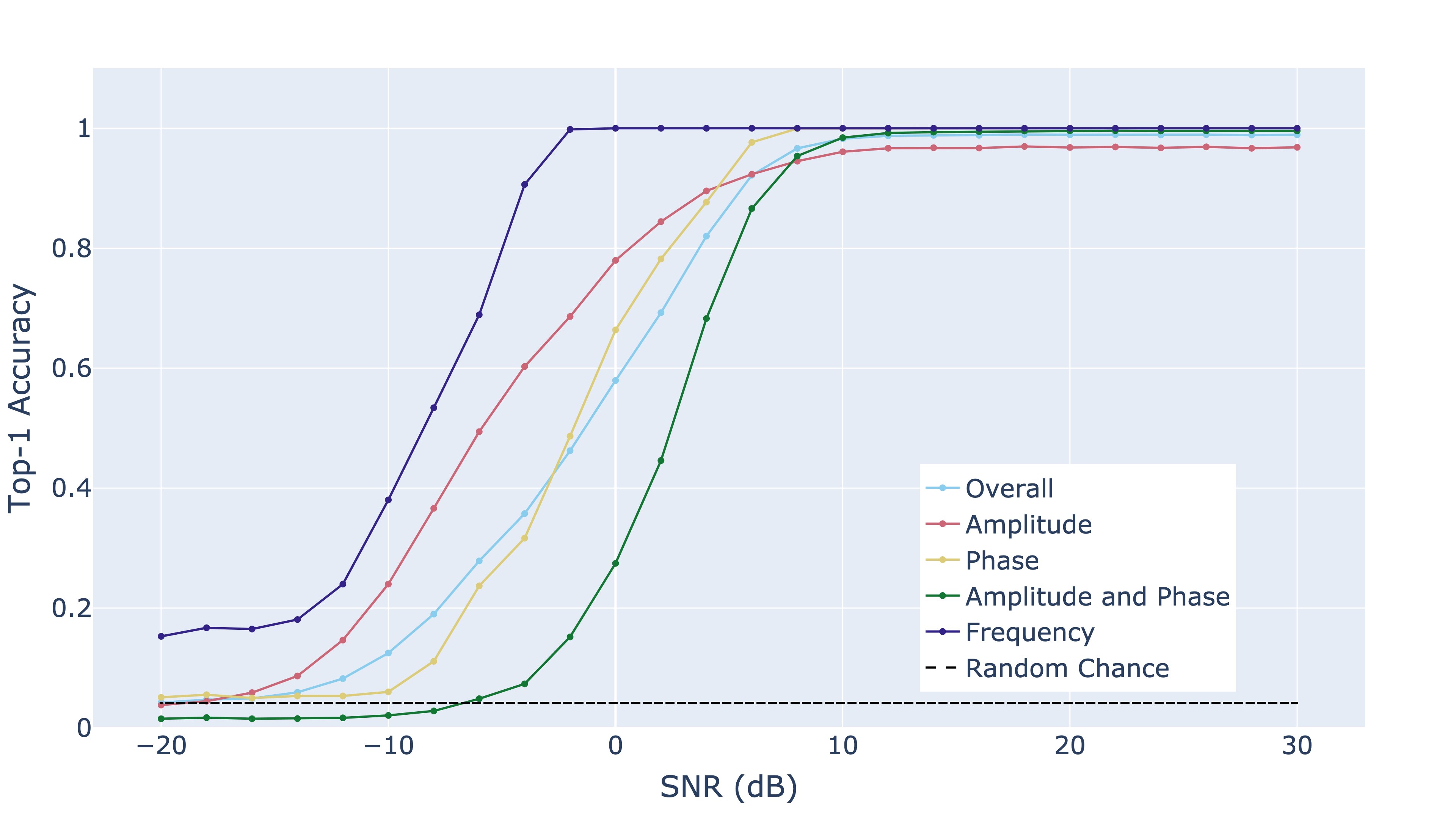}%
    \label{fig::top_1_acc}}
    \hfil
    \subfloat[]{\includegraphics[width=.33\textwidth]{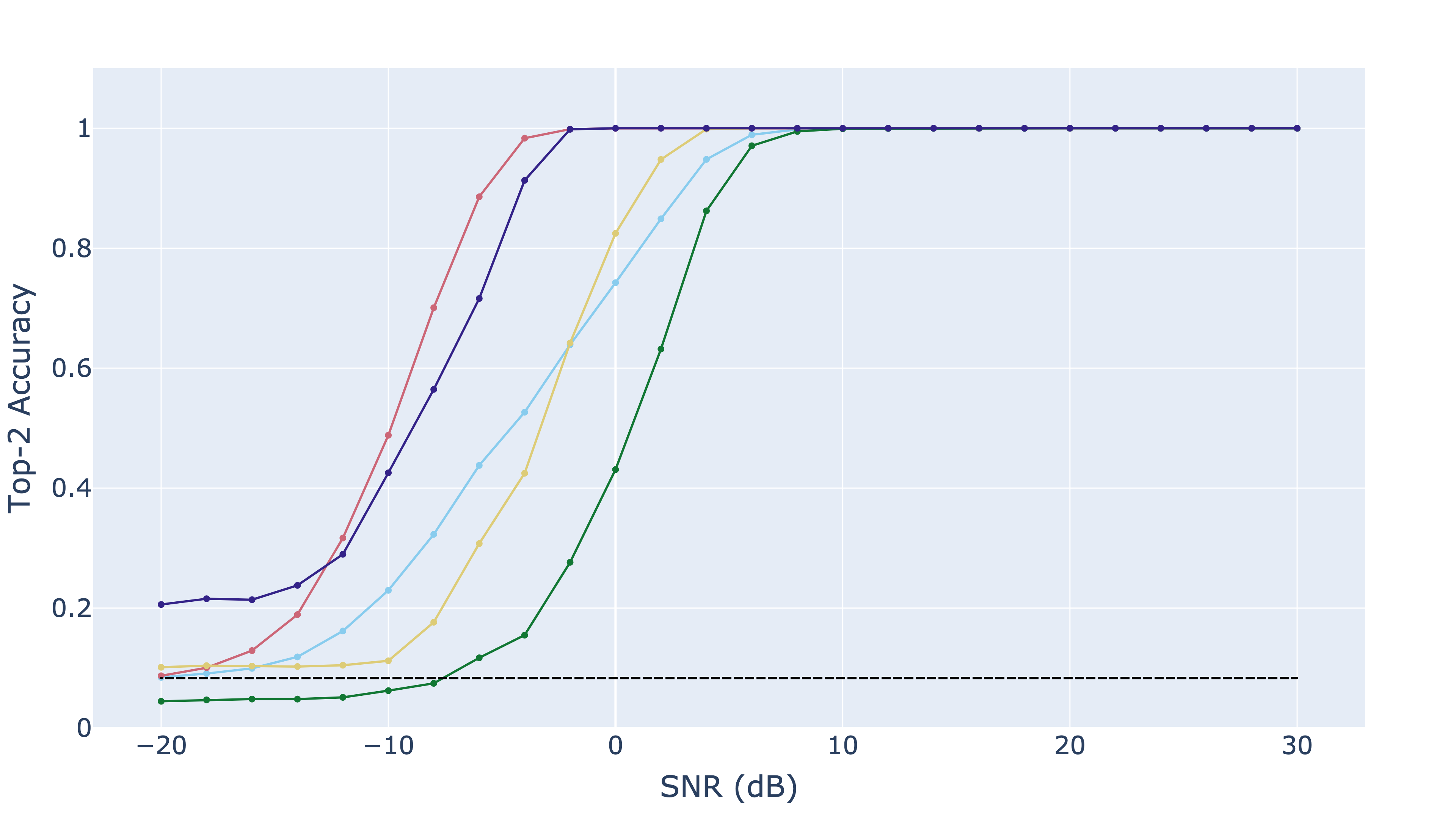}%
    \label{fig::top_2_acc}}
    \subfloat[]{\includegraphics[width=.33\textwidth]{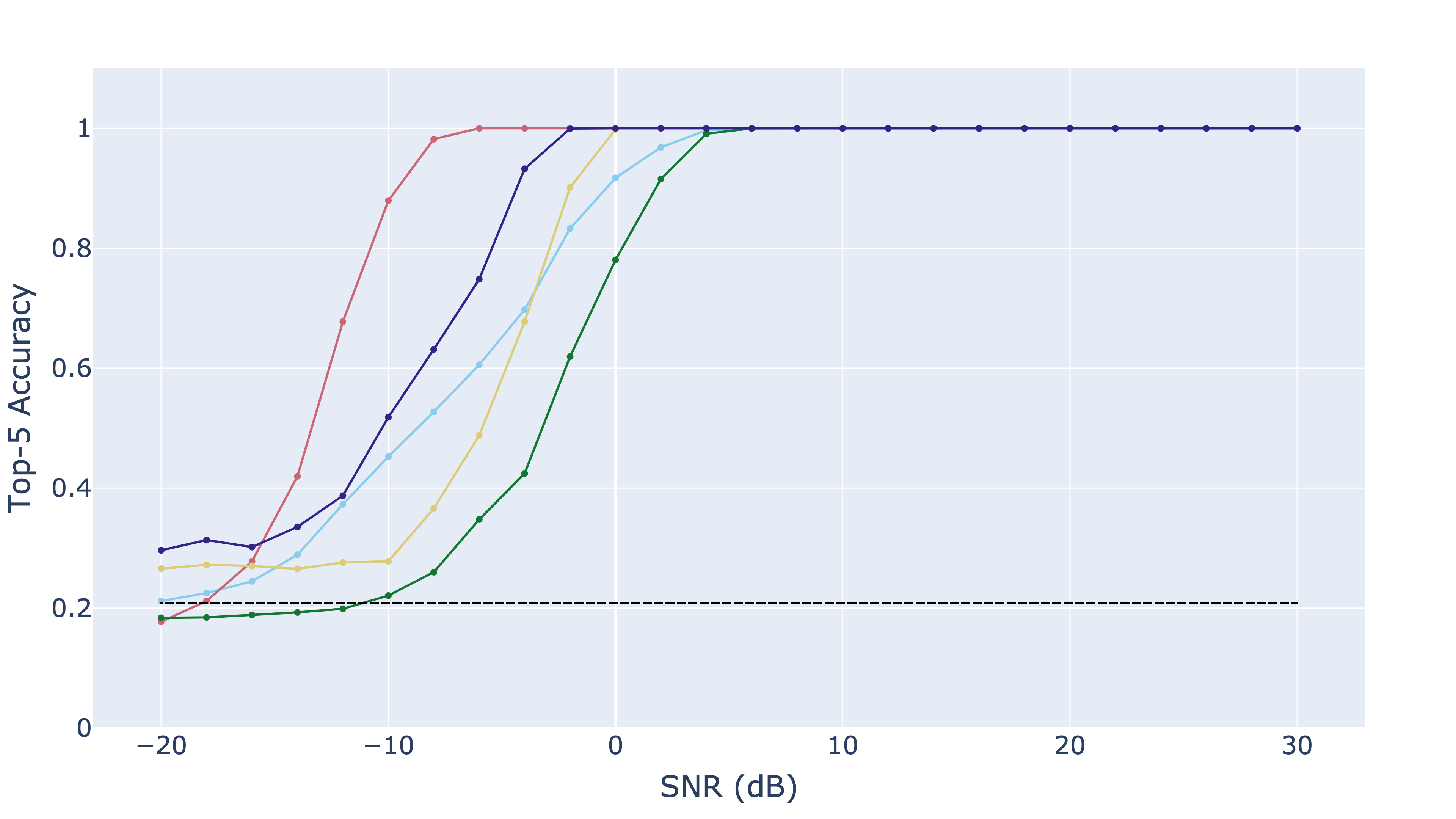}%
    \label{fig::top_5_acc}}
    \caption{Accuracy over varying SNR conditions for model 1110 with (a), (b), and (c) showing the top-1, top-2, and top-5 accuracy respectively. Random chance for each is defined as 1/24, 2/24, and 5/24.}
    \label{fig::top_k_acc}
\end{figure*}

    As discussed, in systems where the modulation schemes must be classified quickly, it is advantageous to apply fewer demodulation schemes in a trial and error fashion.  This is particularly significant at lower SNR values where accuracy is mediocre.  Top-k accuracy allows an in-depth view on the expected number of trials before finding the correct modulation scheme.  Although traditional accuracy (top-1 accuracy) characterizes the performance of the model in terms of classifying the exact variant, top-k accuracy characterizes the percentage of the classifier predicting the correct variant among the top-k predictions (sorted by descending class probabilities).  We plot the top-1, top-2, and top-5 classification accuracy over varying SNR conditions for each modulation grouping defined in \Cref{sec::dataset} in \Cref{fig::top_k_acc}.
    
    Although performance decays to approximately random chance for the overall (all modulation schemes) performance curves for each top-k accuracy, it is notable that some modulation group performances drop below random chance.  The models are trained to maximize the overall model performance.  This could explain why certain modulation groups dip below random chance but the overall performance and other modulation groups remain at or above random chance.
    
    Using the proposed method greatly reduces the correct modulation scheme search space.  While high performance in top-1 accuracy is increasingly difficult to achieve with low SNR signals, top-2 and top-5 accuracy converge to higher values at a much faster rate.  This indicates our proposed method greatly reduces the search space from 24 modulation candidates to fewer candidate types when employing trial and error methods to determine the correct modulation scheme.  Further, if the group of modulation is known (\textit{e.g.}, FM), one can view a more specific tradeoff curve in terms of SNR and top-k accuracy given in \Cref{fig::top_k_acc}.

\subsection{Short Duration Signal Bursts}
    Due to the rapid scanning characteristic of some modern software-defined radios, we investigate the performance tradeoff of varying signal duration and AMC performance. This analysis is meant to emulate the situation wherein a receiver only detects a short RF signal burst.  We investigate signal burst durations of 1.024 ms (full length signal from original dataset), 512 $\mu$s, 256 $\mu$s, 128 $\mu$s, 64 $\mu$s, 32 $\mu$s, and 16 $\mu$s.  We assume the same 1MS/sec sampling rate as in the previous analyses such that 16 $\mu$s burst is captured in 16 I/Q samples.
    
    \begin{figure}[htbp]
    \centering
    \includegraphics[width=\columnwidth]{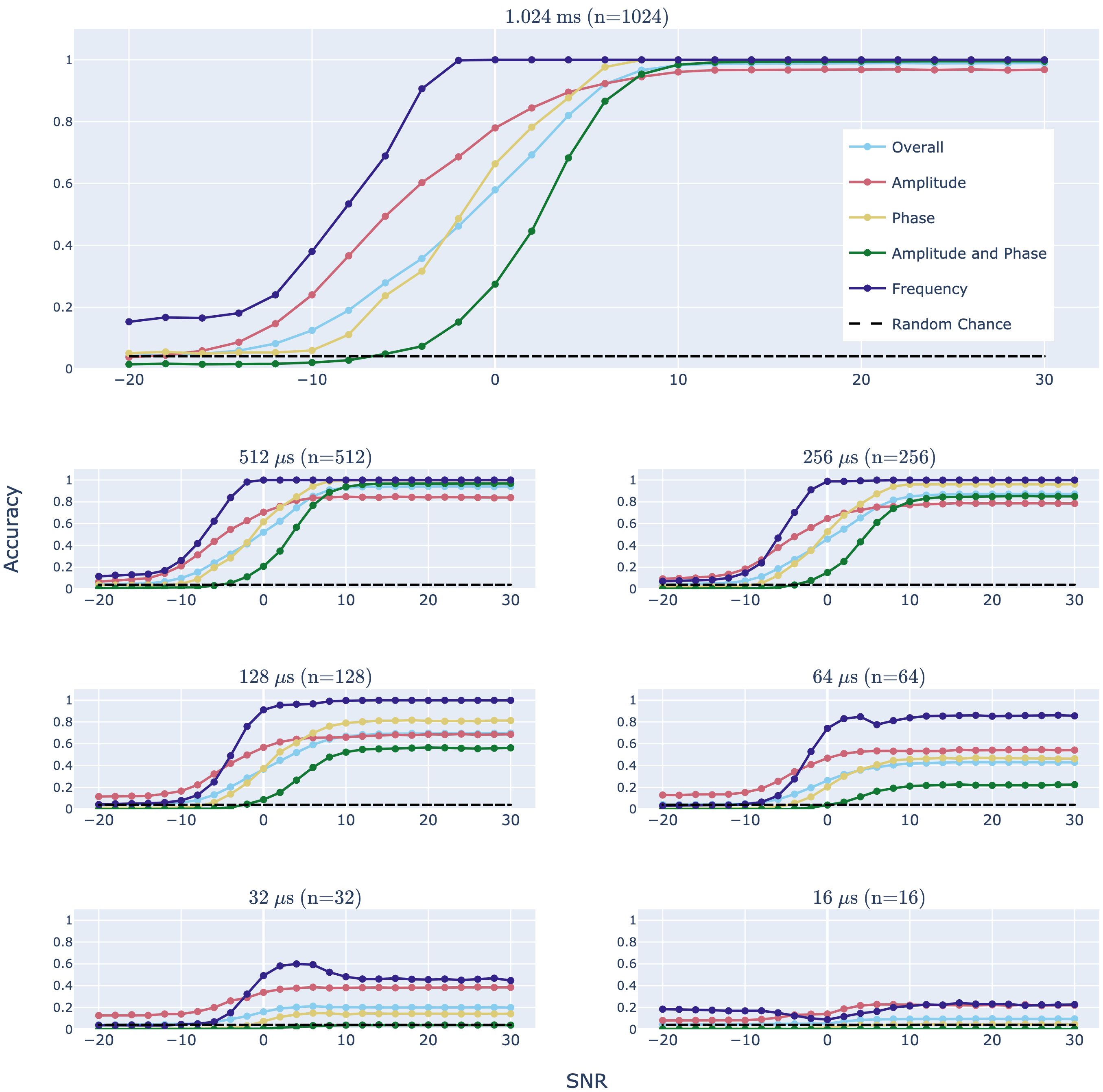}
    \caption{Tradeoff in accuracy for various signal lengths across SNR, grouped by modulation category for the best performing model 1110. The top plot shows the baseline performance using the full sequence. Subsequent plots show the same information using increasingly smaller signal lengths for classification.   }
    \label{fig::lpoi_tradeoff}
    \end{figure} 
    
    In this section, we use the same test set as our other investigations; however, a uniformly random starting point is determined for each signal such that a contiguous sample of the desired duration, starting at the random point, is chosen. Thus, the chosen segment from a test set sample is randomly assigned. 
    
    We also note that, although the sample length for the evaluation is changed, the best performing model is the same architecture with the exact same trained weights because this model uses statistics pooling from the X-Vector inspired modification. A significant benefit to the X-Vector inspired architecture is its ability to handle variable-length inputs without the need of padding, retraining, or other network modifications.  This is achieved by taking global statistics across convolutional channels producing a fixed-length vector, regardless of signal duration.  Due to this flexibility, the same model (model 1110) weights are used for each duration experiment.  This fact also emphasizes the desirability of using X-vector inspired AMC architectures for receivers that are deployed in an environment where short-burst and variable duration signals are anticipated to be present.
    
    \begin{figure*}[!ht]
        \centering
        \subfloat[]{\includegraphics[width=.33\textwidth]{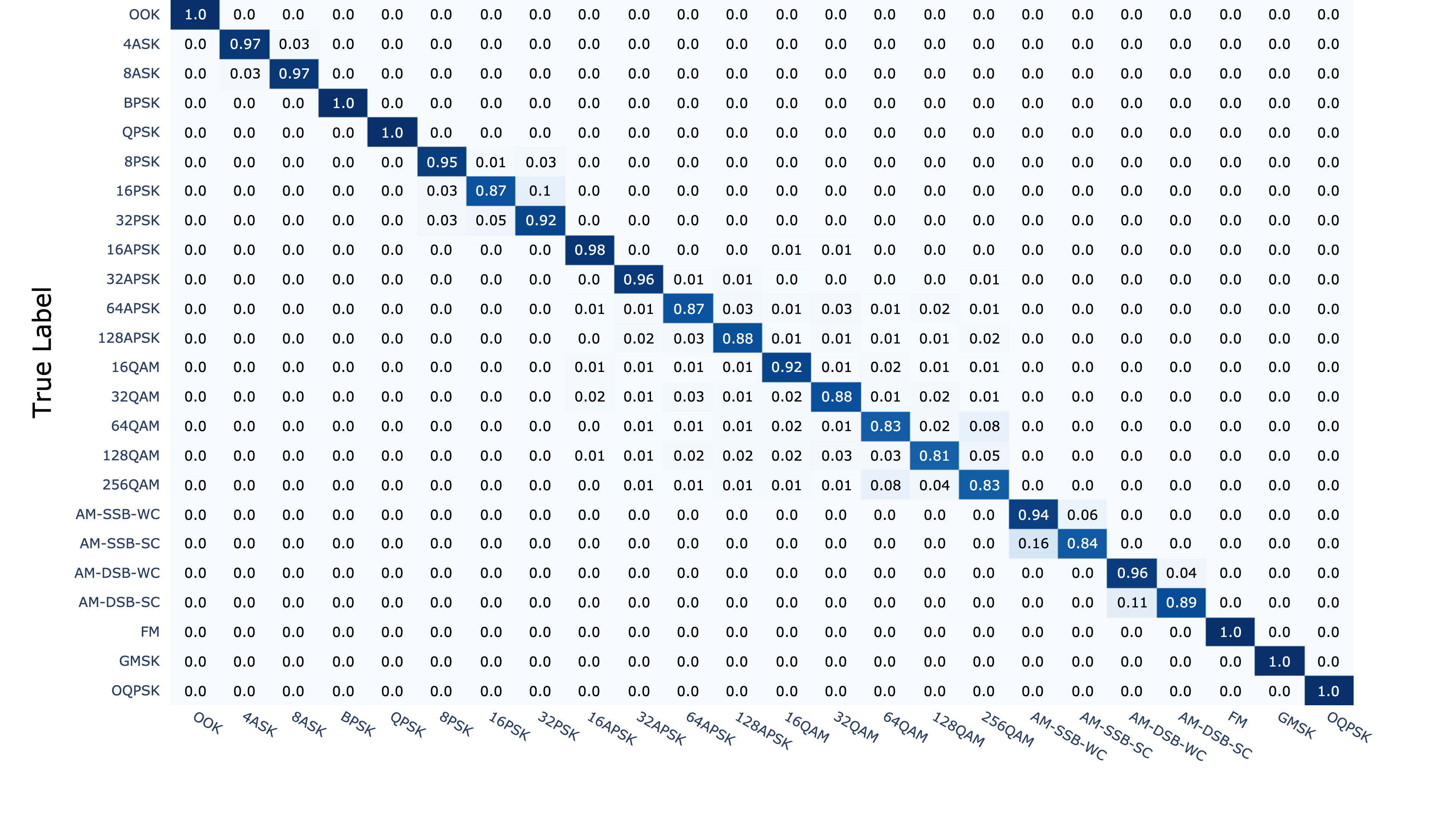}%
        \label{fig::gt_eq_0db}}
        \hfil
        \subfloat[]{\includegraphics[width=.33\textwidth]{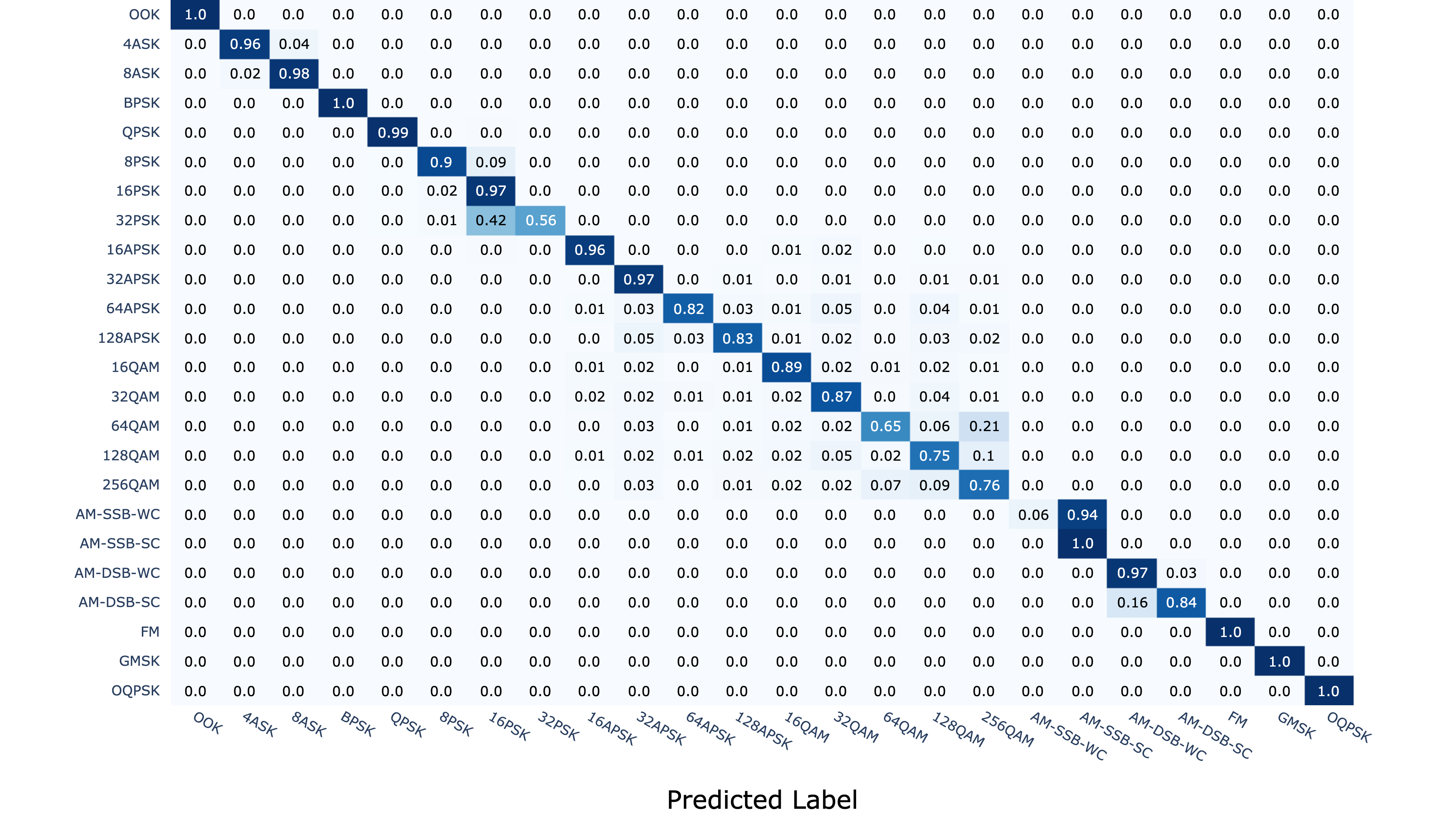}%
        \label{fig::deepsig_geq_0db}}
        \subfloat[]{\includegraphics[width=.33\textwidth]{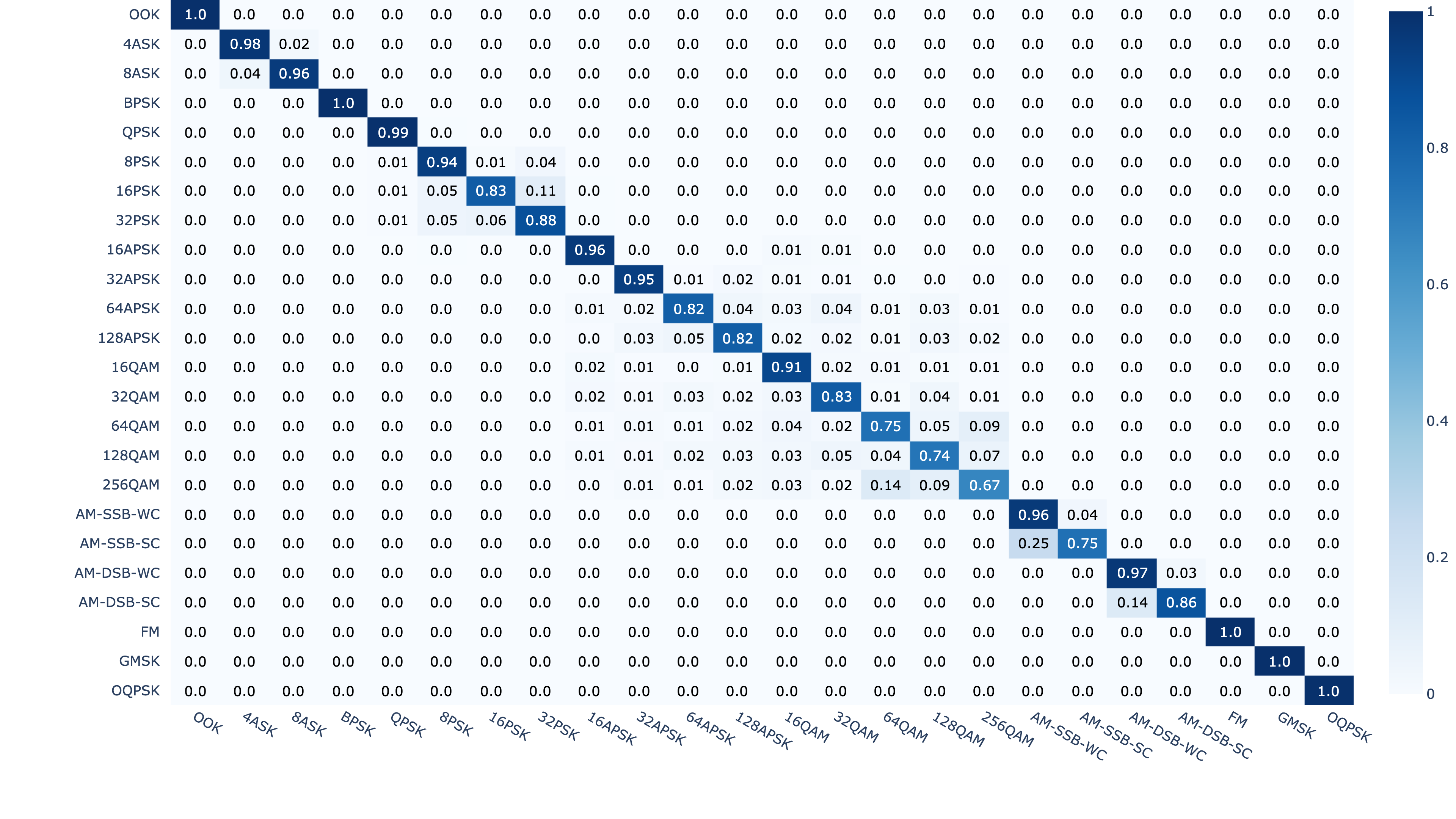}%
        \label{fig::asil_2020_geq_0db}}
        \caption{Confusion matrices for (a) model 1110 (best performing model from this work), (b) the reproduced ResNet model from \cite{deepsig}, and (c) the X-Vector inspired model from \cite{asilomar} with SNR $\geq$ 0dB.}
        \label{fig::confusion_matrices}
    \end{figure*}
    
    For each signal duration in the time domain, we plot the overall classification accuracy over varying SNR conditions as well as the accuracy for each modulation grouping defined in \Cref{sec::dataset}.  \Cref{fig::lpoi_tradeoff} demonstrates the tradeoff for various signal durations where $n$ is the number of samples from the time domain I/Q signal.
    The first observation is, as we would expect, that classification performance degrades with decreased signal duration. For example, the maximum accuracy begins to degrade at 256 $\mu$s and is more noticeable at 128 $\mu$s.  This is likely a result of using sample statistics that result in unstable or biased estimates for short signal lengths since the number of received signal data points are insufficient to characterize the sample statistics used during training.    Random classification accuracy is approximately 4\% and is shown in the black dotted line in \Cref{fig::lpoi_tradeoff}.  Although classification performance decreases with decreased duration, we are still able to achieve significantly higher classification accuracy than random chance down to 16 $\mu$s of signal capture.

    FM (frequency modulation) signals were typically more resilient to noise interference than AM (amplitude modulation) and AM-PM (amplitude and phase modulation) signals in our AMC.  This was observed across all signal burst durations and our top-k accuracy analysis. This behavior indicates that the performance of our AMC for short bursts, in the presence of increasing amounts of noise, is more robust for signals modulated by changes in the carrier frequency and is more sensitive to signals modulated by varying the carrier amplitude.  We attribute this behavior to our AMC architecture, the architecture of the receiver, or a combination of both of the AMC and receiver.

\subsection{Confusion Matrices}

    While classification accuracy provides a holistic view of model performance, it lacks the granularity to investigate where misclassifications are occurring.  Confusion matrices are used to analyze the distribution of classifications for each given class.  For each true label, the proportion of correctly classified samples is calculated along with the proportion of incorrect predictions for each opposing class.  In this way, we can see which classes the model is struggling to distinguish from one another.  A perfect classifier would be the identity matrix where the diagonal values represent the true class matches the predicted class.  Each matrix value represents the percentage of classifications for the true label and each row sums to 1 (100\%).

    \Cref{fig::confusion_matrices} illustrates the class confusion matrices for SNR levels greater than or equal to 0dB for models 1110, the reproduced ResNet architecture from \cite{deepsig}, and the baseline X-Vector architecture from \cite{asilomar_2020} respectively.  Shown in \cite{asilomar_2020}, the X-Vector architecture was able to distinguish PSK and AM-SSB variants to a higher degree and performed better overall than \cite{deepsig}.  Both architectures struggled to differentiate QAM variants.  
    
    Model 1110 improved upon these prior results for QAM signals and in general has higher diagonal components than the other architectures.  This again supports a conclusion that model 1110 achieves a new state-of-the-art in AMC performance.

\section{Conclusion}
    A comprehensive ablation study was carried out with regard to AMC architectural features using the extensive RadioML 2018.01A dataset. This ablation study built upon a strong performance of a new baseline model that was also introduced in the initial investigation of this study. This initial investigation informed the design of a number of AMC architecture modifications---specifically, the use of X-Vectors, dilated convolutions, and SE blocks. With the combined modifications, we achieved a new state-of-the-art in AMC performance. Among these modifications, dilated convolutions were found to be the most critical architectural feature for model performance. Self-attention was also investigated but was not found to increase performance---although increased temporal context improved upon prior works.     

\typeout{} 
\bibliographystyle{IEEEtran}

\bibliography{refs}

\newpage

\begin{IEEEbiography}[{\includegraphics[width=1in,height=1.25in,clip,keepaspectratio]{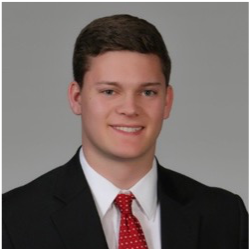}}]{Clayton A. Harper}
received his B.S. in mathematics and computer engineering in 2019 and M.S. in data engineering in 2021 from Southern Methodist University in Dallas, TX, where he specialized in machine learning and audio signal processing. His main research area is the analysis of time series signal processing in computer systems, especially pertaining to security and privacy.  He is a student member of IEEE and is currently pursuing his Ph.D. in computer science at Southern Methodist University with the co-advisors Dr. Eric C. Larson and Dr. Mitchell A. Thornton.
\end{IEEEbiography}

\vspace{11pt}

\begin{IEEEbiography}[{\includegraphics[width=1in,height=1.25in,clip,keepaspectratio]{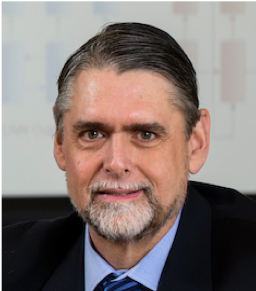}}]{Mitchell (Mitch) A. Thornton}
is currently the Cecil H. Green Chair of Engineering and Professor in the Department of Electrical and Computer Engineering at Southern Methodist University in Dallas, Texas. He also serves as the Executive Director of the Darwin Deason Institute for Cyber Security, a research-only unit, and as Program Director for the interdisciplinary M.S. in Data Engineering degree program within the Lyle School of Engineering at SMU. His main research interests are in the areas of cyber security and quantum informatics.  His past industrial experience includes full-time employment at the Amoco Research Center, E-Systems, Inc. (now L3Harris Technologies Inc.), and the Cyrix Corporation. Dr. Thornton is a member of several professional and honor societies including the IEEE and the ACM where he is a senior member in each organization. He was elected as Chair of the IEEE Technical Community on Multiple-Valued Logic (TCMVL, 2010-11) and has served in various roles for other IEEE/ACM committees. He is an author or co-author of five books and more than 300 technical articles. He is a named inventor on over 20 US/PCT/WIPO patents and patents pending. He holds P.E. licenses in the states of Texas, Mississippi and Arkansas. He received the Ph.D. in computer engineering from SMU in 1995, M.S. in computer science from SMU in 1993, M.S. in electrical engineering from the University of Texas at Arlington in 1990, and B.S. in electrical engineering from Oklahoma State University in 1985.
\end{IEEEbiography}

\vspace{11pt}

\begin{IEEEbiography}[{\includegraphics[width=1in,height=1.25in,clip,keepaspectratio]{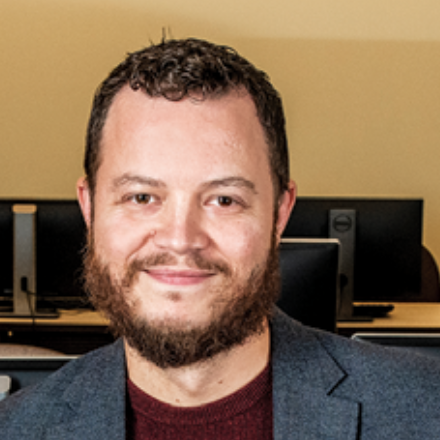}}]{Eric C. Larson}
is an Associate Professor in the department of Computer Science in the Bobby B. Lyle School of Engineering, Southern Methodist University. His main research interests are in machine learning, sensing, and signal / image processing for various applications, in particular, for healthcare and security applications. His work in both areas has been commercialized and he holds a variety of patents for sustainability sensing and mobile phone-based health sensing. Dr. Larson has authored one textbook and over 70 technical articles. He is active in signal processing education for computer scientists and is an active member of IEEE and the ACM. He received his Ph.D. in 2013 from the University of Washington, where he was co-advised by Shwetak N. Patel and Les Atlas. He received his B.S. and M.S. in Electrical Engineering in 2006 and 2008, respectively, at Oklahoma State University, where he was advised by Damon Chandler.
\end{IEEEbiography}

\vfill

\end{document}